\documentclass[lettersize,journal]{IEEEtran}
\usepackage{amsmath,amsfonts}
\usepackage{algorithmic}
\usepackage{algorithm}
\usepackage{array}
\usepackage[caption=false,font=normalsize,labelfont=sf,textfont=sf]{subfig}
\usepackage{textcomp}
\usepackage{stfloats}
\usepackage{url}
\usepackage{verbatim}
\usepackage{graphicx}
\usepackage{cite}
\usepackage{multirow}
\usepackage{booktabs}
\usepackage[colorlinks=true,linkcolor=blue,citecolor=blue]{hyperref}
\usepackage{xcolor}
\usepackage{makecell}
\usepackage{pifont} 
\usepackage{amssymb}
\usepackage{caption}
\definecolor{mplblue}{HTML}{1F77B4}   
\definecolor{mplorange}{HTML}{FF7F0E} 
\definecolor{mplgreen}{HTML}{2CA02C}  
\definecolor{mplred}{HTML}{D62728}    

\author{
Zihan~Zheng,
~Zhenlong~Wu,
~Xuanxuan~Wang,
~Houqiang~Zhong,
~Xiaoyun~Zhang,
~\textit{Member, IEEE},

~Qiang~Hu,
~\textit{Member, IEEE},
~Guangtao~Zhai,~\textit{Fellow, IEEE},
~Wenjun~Zhang,~\textit{Fellow, IEEE}
\thanks{
Zihan Zheng, Zhenlong Wu, Xuanxuan Wang, Houqiang Zhong, Xiaoyun Zhang, Qiang Hu, Guangtao Zhai and Wenjun Zhang are with the Shanghai Jiao Tong University, Shanghai, 200240, China. (email: \{1364406834, 1821863716, wangxuanxuan, zhonghouqiang, xiaoyun.zhang, qiang.hu, zhaiguangtao, zhangwenjun\}@sjtu.edu.cn).

Corresponding author: Qiang Hu (e-mail: qiang.hu@sjtu.edu.cn).
}

}

\title{Constrained Dynamic Gaussian Splatting}

\begin{document}

\markboth{IEEE Transactions On Circuits and Systems For Video Technology,~Vol.~XX, No.~XX, Month~20XX}%
{Zheng \MakeLowercase{\textit{et al.}}: Constrained Dynamic Gaussian Splatting}

\IEEEpubid{%
\parbox[b]{\textwidth}{%
\vspace{1pt}
\scriptsize
Copyright \textcopyright{} 2026 IEEE. Personal use of this material is permitted.
However, permission to use this material for any other purposes must be obtained
from the IEEE by sending an email to pubs-permissions@ieee.org.
}%
}

\maketitle 

\begin{abstract}
While Dynamic Gaussian Splatting enables high-fidelity 4D reconstruction, its deployment is severely hindered by a fundamental dilemma: unconstrained densification leads to excessive memory consumption incompatible with edge devices, whereas heuristic pruning fails to achieve optimal rendering quality under preset Gaussian budgets. 
In this work, we propose Constrained Dynamic Gaussian Splatting (CDGS), a novel framework that  formulates dynamic scene reconstruction as a budget-constrained optimization problem to enforce a strict, user-defined Gaussian budget during training. Our key insight is to introduce a differentiable budget controller as the core optimization driver. Guided by a multi-modal unified importance score, this controller fuses geometric, motion, and perceptual cues for precise capacity regulation. To maximize the utility of this fixed budget, we further introduce an adaptive static-dynamic allocation strategy that separates the Gaussian representation into static and dynamic branches and distributes the shared global capacity between them according to motion complexity. Furthermore, we implement a three-phase training strategy to seamlessly integrate these constraints, ensuring precise adherence to the target count. After training, a dual-mode hybrid compression scheme further reduces storage overhead. CDGS therefore not only strictly adheres to the specified Gaussian-count budget (error \textless \textbf{2\%}) but also achieves favorable rate-distortion performance. Extensive experiments demonstrate that CDGS delivers optimal rendering quality under varying capacity limits and favorable rate-distortion performance, achieving over \textbf{3×} model compression compared with the state-of-the-art method Ex4DGS.
\end{abstract}


\IEEEpeerreviewmaketitle

\section{Introduction}
\label{sec:intro}
Free-viewpoint Video (FVV) enables users to freely and dynamically explore 3D scenes from arbitrary viewpoints, revolutionizing immersive media experiences in VR/AR, sports broadcasting, and telepresence. While offering unprecedented interactivity, the mass deployment of FVV is severely bottlenecked by the heterogeneity of end-user hardware. The strict memory and bandwidth limitations of edge devices stand in sharp contrast to the massive data requirements of high-fidelity volumetric content, making it crucial to maximize reconstruction quality under strict resource constraints.

Early methods for FVV reconstruction typically employed dynamic meshes\cite{HumanMeshRecovery,hsmr}, point clouds~\cite{PointCloudBasedVolumetricVideoCodecs,DynamicPointCloud,graziosi2020overview}, depth maps \cite{fvvdibr,boyce2021mpeg} or image-based view interpolation \cite{varfvv},
frequently resulting in compromised visual quality, especially in complex dynamic scenes. Neural Radiance Fields (NeRF) \cite{mildenhall2021nerf} and its variants \cite{barron2021mip,barron2022mip,chen2022tensorf,muller2022instant,duckworth2024smerf,barron2023zip} marked a breakthrough in novel view synthesis.
Subsequent work on dynamic NeRF \cite{li2022streaming,rerf, videorf, zheng2024jointrf, zheng2024hpc} extensions incorporating temporal modeling further expanded FVV's potential. However, practical deployment remains challenging due to slow rendering, inconsistent output quality, and substantial computational requirements.

\begin{figure*}
\centering

    \includegraphics[width=\linewidth]{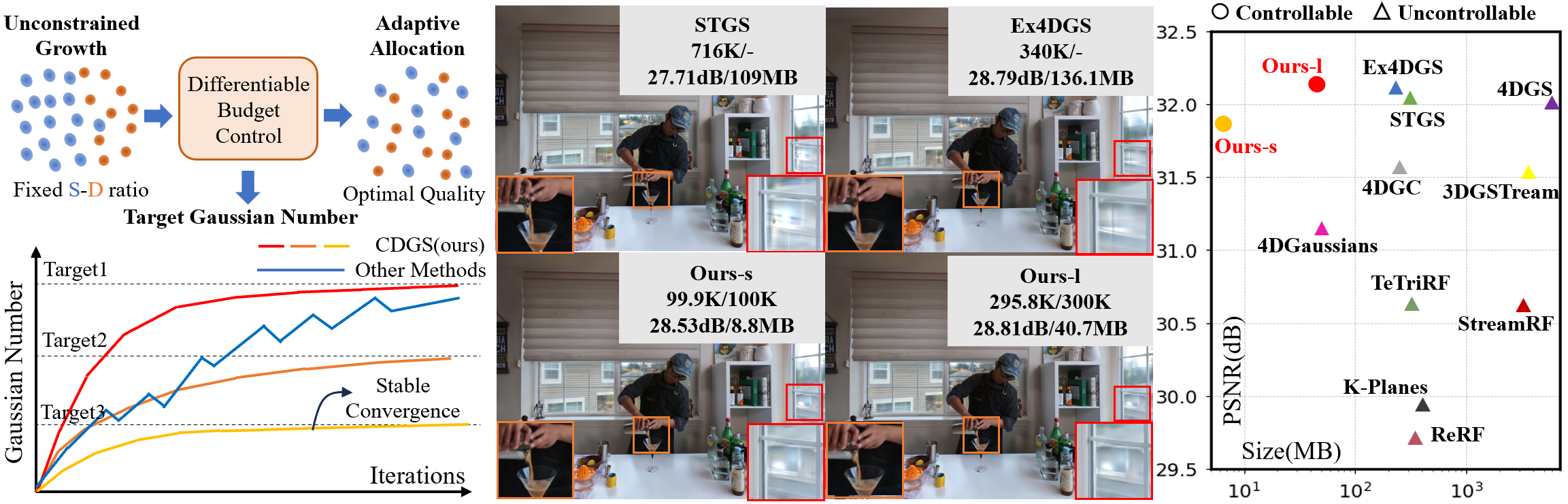}

    \captionof{figure}{
      \textbf{Left}: Our CDGS leverages differentiable budget control for precise Gaussian number regulation, achieving adaptive static-dynamic allocation across varying target numbers and optimal rendering quality. \textbf{Middle}: Visual comparison with state-of-the-art methods, highlighting advantages in visual quality, model size, and Gaussian count controllability (second row: actual/target counts). \textbf{Right}: Superior rate-distortion performance and precise Gaussian number control of our approach, outperforming the compared prior works (e.g., 4DGS \cite{yang2023gs4d}, STGS \cite{Li_STG_2024_CVPR}, and Ex4DGS \cite{lee2024ex4dgs}).
    }
    \label{fig:teaser}

\end{figure*}


Recently, 3D Gaussian Splatting (3DGS) \cite{kerbl20233d} has revolutionized real-time rendering. However, dynamic extensions \cite{Li_STG_2024_CVPR,lee2024ex4dgs,Wu_2024_CVPR,yang2023gs4d} typically struggle with explosive growth in Gaussian counts, where unconstrained densification defies hardware limits. While recent advancements attempt to address efficiency, they fail to simultaneously achieve controllability and optimal allocation. For instance, Ex4DGS \cite{lee2024ex4dgs} reduces redundancy via decomposition but lacks explicit capacity control, relying on heuristics that yield unpredictable model complexity. Conversely, Taming 3DGS \cite{10.1145/3680528.3687694} controls the Gaussian population during training by scheduling discrete densification and pruning operations under a prescribed resource budget, but it is designed for static scenes and does not formulate the target count as a differentiable loss. Applying such a strategy directly to dynamic settings remains inadequate because static-scene importance criteria do not account for kinematic significance or the competition for capacity between static and dynamic representations. Our key insight is that constrained dynamic reconstruction requires integrating a differentiable budget objective into the training loop while adaptively balancing resources between static and dynamic components.

In this paper, we propose Constrained Dynamic Gaussian Splatting (CDGS), a framework that reformulates dynamic scene reconstruction as a budget-constrained optimization problem, as illustrated in Fig. \ref{fig:teaser}. Departing from the conventional paradigm of uncontrolled growth, CDGS treats the Gaussian count as a strict budget. By optimizing the spatio-temporal distribution within this user-defined limit, it not only ensures precise controllability but also pushes the Pareto frontier of rate-distortion performance. Our approach is built upon four key components. First, to enforce the target capacity, we introduce a differentiable budget controller. This mechanism is driven by a differentiable budget loss guided by a multi-modal unified importance score, which fuses geometric stability, kinematic significance, and perceptual impact. This ensures that visually critical dynamic details are preserved even under tight budgets.

\IEEEpubidadjcol

Second, to ensure robust adaptability across varying scene dynamics, we introduce an adaptive static-dynamic allocation method. Instead of relying on heuristic ratios, we analyze the motion distribution to automatically determine a scene-adaptive separation threshold, allowing the relative static-dynamic composition to emerge under the shared global budget. Third, we implement a three-phase training strategy to seamlessly integrate these constraints, ensuring precise adherence to the target count (error\textless\textbf{2\%}). Finally, to minimize storage footprint, a dual-mode hybrid compression strategy is tailored specifically for the decomposed static and dynamic streams.

Collectively, these innovations enable CDGS to deliver controllable, compact, and high-quality FVV representations tailored to arbitrary hardware specifications. Experimental results demonstrate favorable rate-distortion performance, achieving over \textbf{3×} model compression compared with the state-of-the-art method Ex4DGS. In summary, our contributions are as follows:

\begin{itemize}
\item 
We reinterpret dynamic Gaussian splatting as a budget-constrained optimization problem, enabling controllable model complexity and predictable capacity. 

\item 
We introduce a differentiable budget controller guided by a unified importance score, together with an autonomous adaptive static-dynamic allocation strategy  that optimizes Gaussian distribution under a fixed budget.

\item 
We design a budget-consistent three-phase training scheme that enforces strict budget adherence, together with a post-training dual-mode hybrid compression pipeline that minimizes spatio-temporal redundancy.

\item 
Extensive experiments across multiple datasets show that CDGS consistently outperforms existing dynamic scene reconstruction methods, achieving superior rate-distortion performance and precise Gaussian count control.

\end{itemize}


\section{Related Work}
\label{sec:related work}
\begin{figure*}[t]
\centering
\includegraphics[width=\linewidth]{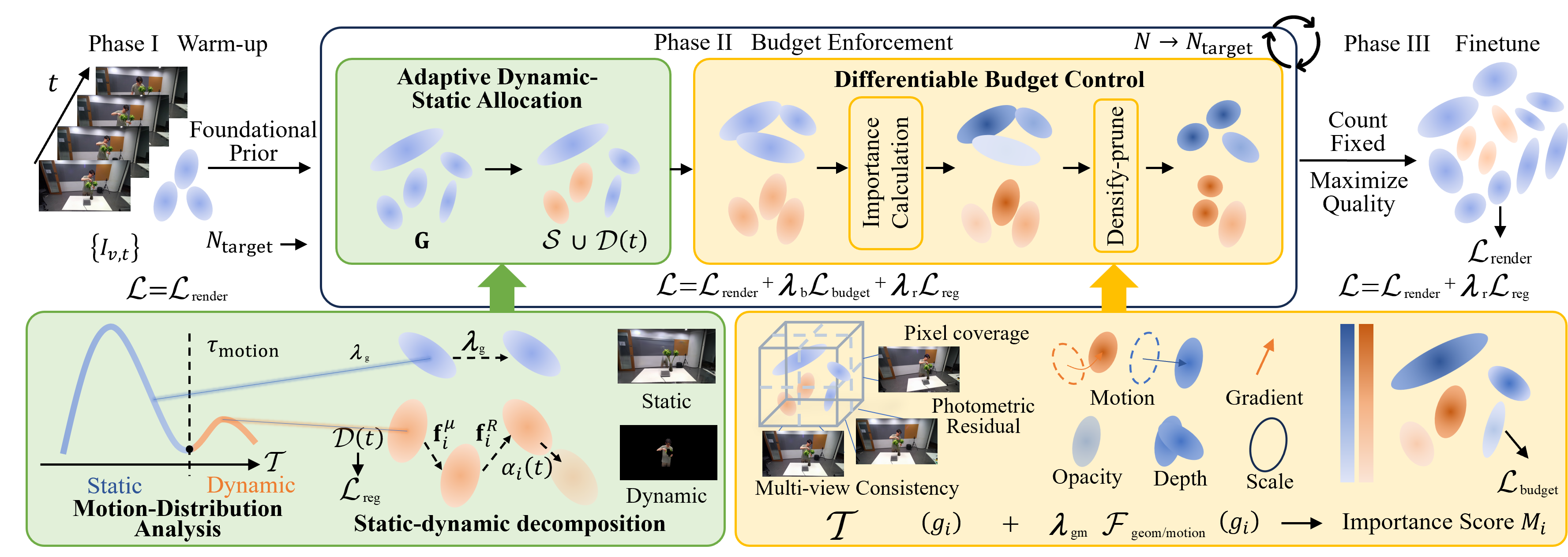}

\caption{
Overview of the proposed CDGS framework. (\textbf{Top}) Three-Phase Pipeline: The training progresses from a Warm-up phase to establish foundational priors, through a Budget Enforcement phase where constraints are actively applied, to a final Fine-tuning phase that maximizes quality under the fixed count $N_{\text{target}}$. (\textbf{Bottom Left}) Adaptive Static-Dynamic Allocation: This module analyzes the distribution of motion magnitudes to identify a distribution-adaptive separation threshold $\tau_{\text{motion}}$. It decomposes the scene into a Static set $\mathcal{S}$ and a Dynamic set $\mathcal{D}(t)$. (\textbf{Bottom Right}) Differentiable Budget Control: This module regulates capacity by computing a Unified Importance Score $M_i$. It fuses Perceptual cues $\mathcal{F}_{\text{perceptual}}$ and Geometric/Motion cues $\mathcal{F}_{\text{geom/motion}}$. The resulting score guides the differentiable budget loss $\mathcal{L}_{\text{budget}}$ to precisely prune or densify Gaussians towards $N_{\text{target}}$.
}
\label{fig:Method}

\end{figure*}
\subsection{Novel View Synthesis for Static Scenes}


NeRF \cite{mildenhall2021nerf} revolutionized novel view synthesis by parameterizing continuous radiance fields with Multi-Layer Perceptrons, yet its reliance on expensive ray marching imposes prohibitive computational costs. To mitigate this, subsequent works \cite{barron2021mip,zhang2020nerf++,barron2022mip,chen2022tensorf,muller2022instant,sun2022direct,karnewar2022relu,niemeyer2025radsplat,zhang2022nerfusion,duckworth2024smerf,barron2023zip} adopted structured representations such as voxel grids, octrees, and hash tables to accelerate the query process. While these hybrid approaches significantly improve training and rendering efficiency, they fundamentally remain bound by the ray-marching paradigm, where the requisite point-wise sampling along each ray continues to limit the optimal balance between high-frequency detail capture and computational throughput.

Marking a paradigm shift from implicit ray marching, 3DGS \cite{kerbl20233d} and its variants \cite{charatan2024pixelsplat,feng2025flashgs,gao20246dgs,hollein20243dgs, navaneet2023compact3d,fan2023lightgaussian,scaffoldgs,wang2024contextgs,zhong20254dmodeeditablescalablevolumetric,gao2025aligngsaligninggeometrysemantics} leverage explicit anisotropic Gaussians and tile-based GPU rasterization to achieve superior visual quality in real-time. Compression-oriented methods such as EAGLES \cite{girish2024eaglesefficientaccelerated3d}, C3DGS \cite{lee2024c3dgs}, Compressed 3D Gaussian Splatting \cite{10655416}, and GSCodec Studio \cite{11355359} reduce the storage cost of Gaussian representations through lightweight encoding, compact attributes, quantization, or modular coding tools. However, the adaptive density control strategy of 3DGS still allows the number of primitives to grow unboundedly, resulting in high and variable memory usage that often exceeds consumer-grade hardware capacities. This uncontrolled model complexity and unstable convergence pattern severely hinder deployment under strict computational or memory budgets, necessitating more robust control mechanisms.

\subsection{Novel View Synthesis for Dynamic Scenes}


Extending view synthesis to dynamic environments requires addressing complex motion and temporal redundancy. Early approaches adapted the NeRF framework, either by introducing deformation fields \cite{du2021neural,li2022streaming,park2021nerfies,pumarola2021d,song2023nerfplayer,rerf, videorf, zheng2024jointrf, zheng2024hpc} to map time-variant observations to a canonical space, or by adopting 4D spatio-temporal representations \cite{cao2023hexplane,fang2022fast,fridovich2023k,icsik2023humanrf,li2022neural,park2023temporal,shao2023tensor4d,wang2023mixed,wang2022fourier} like 4D grids. While deformation-based methods struggle with large topological changes, 4D spatio-temporal representations often incur significant memory costs and inherit the slow rendering speeds of implicit methods.

With the advent of 3DGS, research has shifted towards explicit dynamic modeling, generally falling into two streams. The first uses unified 4D primitives, deformation fields, or time-dependent attributes for continuous reconstruction \cite{Li_STG_2024_CVPR,Wu_2024_CVPR,yang2023deformable3dgs,yan20244d,yang2023gs4d,lee2024ex4dgs,wang2025freetimegs,zhang2024mega,cho20254dscaffoldgaussiansplatting,jinmoe,yilmaz2026graphical,lee2025rd4dgs,li2025gifstream4dgaussianbasedimmersive,wu2025swift4d}, including recent motion-aware, deformable, compact, and adaptive representations. The second employs frame-by-frame tracking or streaming updates to handle temporal changes \cite{luiten2023dynamic,sun20243dgstream,hu20254dgc,zheng20254dgcpro,hicom2024,zhong2025prismgsphysicallygroundedantialiasinghighfidelity,fu2025neurips-recongs,ke2025edgegaussian}. Recent TCSVT studies represent complementary directions: Motion-Aware 3DGS \cite{10757420} and FRPGS \cite{10947553} primarily focus on dynamic modeling and reconstruction efficiency, while Light4GS \cite{Light4GS} targets compact 4D Gaussian representations. GSCodec Studio \cite{11355359}, discussed above, provides a modular framework for Gaussian representation compression. Methods such as Ex4DGS \cite{lee2024ex4dgs} and DeGauss \cite{wang2025degaussdynamicstaticdecompositiongaussian} additionally decompose scenes into static and dynamic components. In contrast, CDGS directly optimizes dynamic reconstruction under a user-specified global Gaussian budget and adaptively determines the relative static-dynamic composition under the shared capacity.

\subsection{Gaussian Number Management}


Deploying 3DGS on resource-constrained devices faces a fundamental conflict between limited hardware capacities and the standard optimization strategy, which minimizes error via unbounded densification. Importance-based pruning methods such as LightGaussian \cite{fan2023lightgaussian} reduce redundant Gaussian populations after or near the end of unconstrained optimization. However, such train-then-prune strategies do not directly optimize reconstruction under a prescribed target count, and the remaining Gaussians are not jointly optimized throughout training under the final capacity constraint.

The most closely related population-control methods are Taming 3DGS \cite{10.1145/3680528.3687694} and 3DGS-MCMC \cite{kheradmand20253dgaussiansplattingmarkov}. Taming 3DGS introduces a controlled growth schedule to maintain the Gaussian count near a target level, while 3DGS-MCMC supports controllable population optimization through stochastic relocation. Both focus on static scenes and therefore do not address temporal redundancy, motion-dependent importance, or the allocation of capacity between static and dynamic representations. To bridge this gap, CDGS integrates the target count directly into dynamic-scene training and combines a differentiable budget controller with adaptive static-dynamic allocation, seeking the optimal representation under the specified global capacity.

\section{Method}
\label{sec:3}

We propose CDGS, a unified framework designed for high-fidelity dynamic 3D scene reconstruction under explicit resource constraints. Unlike prior methods prone to heuristic pruning or uncontrolled Gaussian growth, CDGS reformulates the reconstruction task as a budget-constrained optimization problem, enabling precise control over model capacity. As illustrated in Fig.~\ref{fig:Method}, given multi-view video inputs and a target Gaussian count $N_{\text{target}}$, our pipeline is driven by a Differentiable Budget Controller (Sec.~\ref{sec:budget}). This module supervises densification and pruning based on a unified importance score that integrates geometric, motion, and perceptual cues. To maximize the utility of the fixed budget, an Adaptive Static-Dynamic Allocation module (Sec.~\ref{sec:alloc}) determines the relative static-dynamic composition under the shared global capacity. The overall training follows a three-phase pipeline (Sec.~\ref{sec:training}) to stabilize the representation, and employs a dual-mode hybrid compression scheme (Sec.~\ref{sec:compress}) to minimize storage and transmission overhead for efficient deployment.

\subsection{Problem Formulation}
\label{sec:3.1}
Given calibrated multi-view dynamic frames $\{I_{v,t}\}$, where $v$ indexes views $V$ and $t\in\{1,\ldots,T\}$ indexes the $T$ timestamps, 
our goal is to reconstruct a compact spatio-temporal Gaussian scene representation $\mathbf{G}$ 
that enables real-time rendering under a fixed capacity budget $N_{\text{target}}$.
Following 3D Gaussian Splatting \cite{kerbl20233d}, the scene is modeled as a set of anisotropic Gaussians $\mathbf{G}=\{g_i\}$:
\begin{equation}
g_i = \{\boldsymbol{\mu}_i, \mathbf{R}_i, \mathbf{s}_i, \alpha_i, \mathbf{f}_i\},
\end{equation}
where $\boldsymbol{\mu}_i$ denotes the 3D center position. The geometric shape of each Gaussian is determined by a 3D covariance matrix $\Sigma_i$, which is decomposed into a rotation matrix $\mathbf{R}_i$ and a scaling matrix $\mathbf{s}_i$ to ensure positive semi-definiteness during optimization:
\begin{equation}\Sigma_i = \mathbf{R}_i \mathbf{s}_i \mathbf{s}_i^T \mathbf{R}_i^T.\end{equation}

In terms of appearance, each Gaussian carries an opacity scalar $\alpha_i \in [0, 1]$ and a set of spherical harmonics (SH) coefficients $\mathbf{f}_i$ which encode view-dependent radiance. To render novel views, the 3D Gaussians are projected onto the 2D image plane using splatting techniques. The final color $\hat{I}_{v,t}$ for a specific view $v$ at time $t$ is computed via differentiable $\alpha$-blending of the sorted Gaussians overlapping a pixel:
\begin{equation}
\label{eq:render}
\hat{I}_{v,t} = \sum_{i=1}^{|\mathbf{G}|} \text{color}_{i,v} \alpha'_i \prod_{j=1}^{i-1} (1 - \alpha'_j),
\end{equation}
where $\text{color}_{i,v}$ is the view-dependent color decoded from $\mathbf{f}_i$ for view $v$, and $\alpha'_i$ represents the effective opacity in 2D projection space. To drive the optimization of the Gaussian parameters, we minimize the discrepancy between the rendered image $\hat{I}_{v,t}$ and the corresponding ground truth $I_{v,t}$. We supervise the reconstruction with a perceptual appearance loss:
\begin{equation}
\mathcal{L}_{\text{render}}
= (1 - \lambda_{\text{ssim}})\|\hat{I}_{v,t} - I_{v,t}\|_1
+ \lambda_{\text{ssim}}\, \mathcal{L}_{\text{SSIM}}(\hat{I}_{v,t}, I_{v,t}),
\label{rendering_loss}
\end{equation}  

To make the temporal representation explicit before introducing the budget controller, we decompose the full set into two disjoint branches,
\begin{equation*}
\mathbf{G}=\mathcal{S}\cup\mathcal{D}(t),\qquad
\mathcal{S}\cap\mathcal{D}(t)=\varnothing.
\end{equation*}
Here, $\mathcal{S}$ denotes the static branch used for low-motion and regular-motion content. Each Gaussian in this branch retains a lightweight per-Gaussian translational attribute $\mathcal{T}_i$, which combines with the timestamp $t$ to determine its time-dependent position. The dynamic branch $\mathcal{D}(t)$ uses explicit position and rotation trajectories $\mathbf{f}^{\mu}_i$ and $\mathbf{f}^{R}_i$, together with temporal activation parameters, to represent more complex motion. The membership of these two branches is inferred and periodically updated by the adaptive allocation mechanism in Sec.~\ref{sec:alloc}.

Unlike prior dynamic Gaussian approaches that freely grow and prune Gaussians post-training, 
we explicitly constrain the representational capacity during optimization:
\begin{equation}
\label{eq:core_obj}
\min_{\mathbf{G}} \; \mathcal{L}_{\text{render}}(\mathbf{G})
\quad \text{s.t.} \quad |\mathbf{G}| \le N_{\text{target}}.
\end{equation}

This constraint directly governs runtime memory, rendering cost, and even streaming bitrate. 
Our objective is therefore not the unconstrained best reconstruction, but the best reconstruction achievable under a fixed Gaussian-number budget, which forms the foundation for our differentiable population control and adaptive allocation strategies described next.

\subsection{Differentiable Budget Control}
\label{sec:budget}

\begin{figure}[t]
\centering
\includegraphics[width=\linewidth]{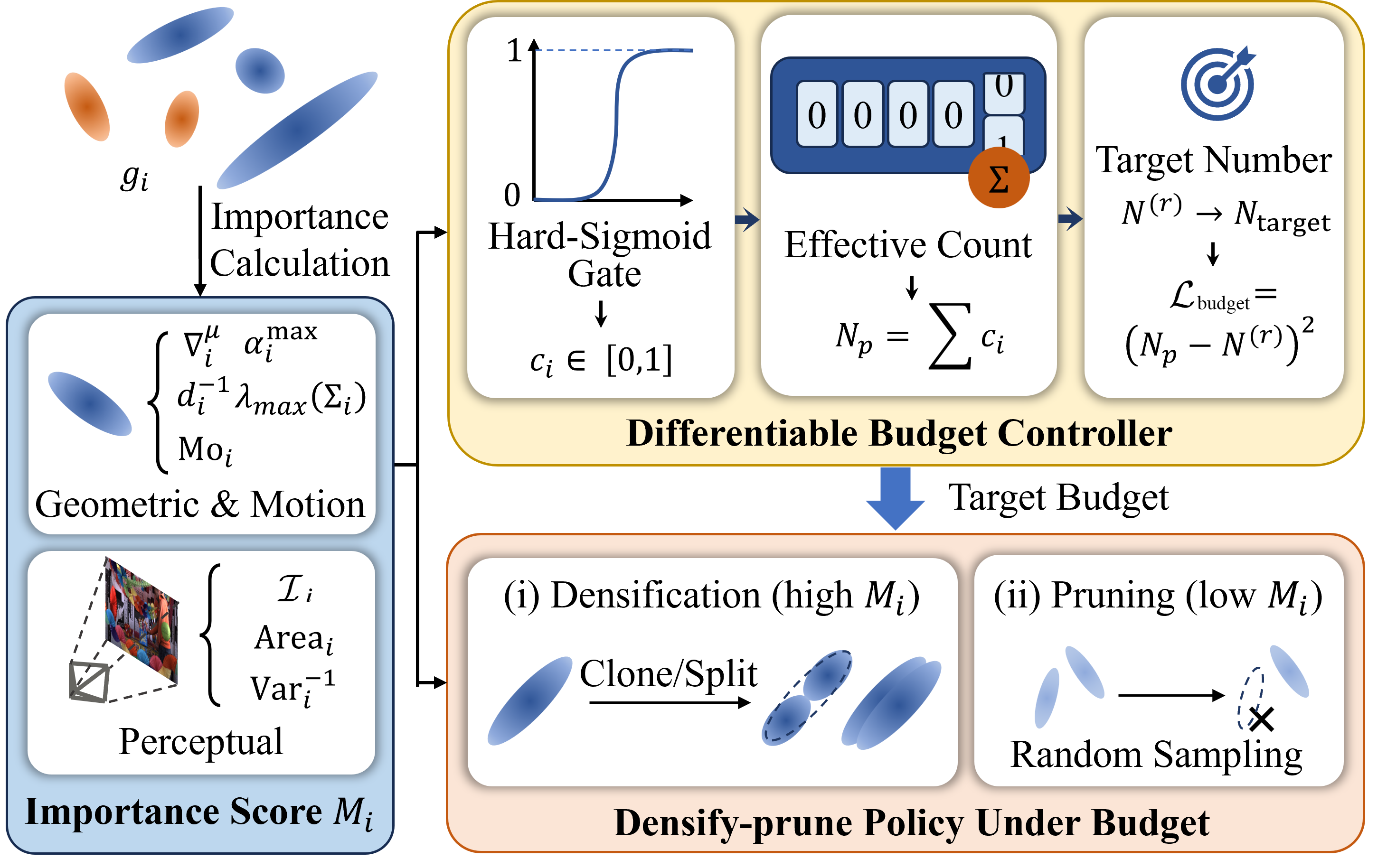}

\caption{Illustration of the Differentiable Budget Controller. The controller aggregates geometric, motion, and perceptual cues to compute a Unified Importance Score $M_i$. This score passes through a hard-sigmoid gate to estimate the effective Gaussian count $N_p$, which is regulated toward the current global sub-target $N(r)$ via a quadratic budget loss $\mathcal{L}_{\text{budget}}$. Guided by this progressively updated constraint, the closed-loop policy performs densification on high-importance Gaussians and pruning on low-importance ones until the final target $N_{\text{target}}$ is reached.
}
\label{fig:budget}

\end{figure}

Enforcing the hard constraint $|\mathbf{G}| \le N_{\text{target}}$ in Eq.~\ref{eq:core_obj} is challenging, 
since the Gaussian count is discrete and non-differentiable. 
We therefore design a differentiable population controller, as illustrated in Fig. \ref{fig:budget}. The controller first computes a Unified Importance Score $M_i$ from geometric, motion, and perceptual cues, and then uses this score to (1) regulate the contribution of Gaussians in rendering, (2) guide adaptive densification and pruning, and (3) penalize deviations from the target capacity through a differentiable budget loss. The detailed construction of $M_i$ is introduced after the differentiable counting mechanism.

\textbf{Differentiable Counting.}
Each Gaussian $g_i$ is assigned a continuous activation variable $c_i \in [0,1]$, 
obtained by applying a temperature-controlled hard-sigmoid gate to its computed importance score $M_i$. 
Before budget enforcement is activated, rendering uses the standard projected opacity $\alpha'_i$ during warm-up, which is equivalent to setting $c_i=1$.
During rendering, $c_i$ directly regulates the participation of the Gaussian,
so Gaussians with $c_i \approx 0$ contribute negligibly. 
The effective active count is estimated as a differentiable proxy:
\begin{equation}
N_p = \sum_i c_i.
\end{equation}
Rather than imposing the final target immediately, we progressively update a global sub-target through a quadratic schedule. Let $N_0$ denote the Gaussian count at the beginning of budget enforcement, $N_T=N_{\text{target}}$ the final target, and $K$ the total number of population-update steps. At the $r$-th update,
\begin{equation*}
N(r)=\left\lfloor ar^2+br+N_0\right\rfloor,
\end{equation*}
where
\begin{equation*}
b=\frac{2(N_T-N_0)}{K},\qquad
a=\frac{N_T-N_0-bK}{K^2}.
\end{equation*}
The current sub-target is used in the quadratic budget loss:
\begin{equation}
\label{eq:budget_loss}
\mathcal{L}_{\text{budget}}^{(r)} = \left(N_p-N(r)\right)^2,
\end{equation}
which drives $N_p$ progressively toward the user-specified final target. The schedule advances at each population-update step during the budget-enforcement stage.
The activation variable is computed as:
\begin{equation}
c_i = \text{clamp}\left( \frac{M_i - 0.5}{\tau_c} + 0.5, \ 0, \ 1 \right),
\end{equation}
where $\tau_c$ is the temperature parameter controlling the steepness of the gate.

To physically enforce this selection, once budget enforcement begins, the opacity used in training-time rendering changes from $\alpha'_i$ to the gated opacity $\hat{\alpha}_i$, defined as:
\begin{equation}
\hat{\alpha}_i = c_i \cdot \alpha'_i.
\end{equation}

This modulation serves as a differentiable bridge: when $c_i\to 0$, the Gaussian becomes transparent and ceases to contribute to the image, physically aligning the soft deletion with the visual output. The rendering loss $\mathcal{L}_{\text{render}}$ (Eq. \ref{rendering_loss}) is therefore coupled to the gating process through $c_i$, while Gaussians with high importance scores are protected from pruning.

The gate temperature is gradually annealed so that $c_i$ approaches a binary mask. Specifically, the temperature $\tau_c$ decays exponentially from an initial value $\tau_{\text{init}}$ to a final value $\tau_{\text{end}}$ during the budget enforcement phase. The decay schedule is formulated as:
\begin{equation}
\tau_c(k) = \tau_{\text{init}} \cdot \left( \frac{\tau_{\text{end}}}{\tau_{\text{init}}} \right)^{\frac{k - k_{\text{start}}}{k_{\text{end}}-k_{\text{start}}}},
\end{equation}
where $k$ denotes the current iteration, and $[k_{\text{start}}, k_{\text{end}}]$ represents the duration of the enforcement phase. At the end of the budget-enforcement phase, thresholding $c_i$ yields an explicit active set with size $\approx N_{\text{target}}$. This converts the discrete constraint into a differentiable penalty relaxation that is fully compatible with gradient-based optimization.

\textbf{Unified Importance Score.}
To determine which Gaussians should survive under a strictly constrained budget, we require a metric that evaluates the contribution of each primitive to the final reconstruction. While resource-aware methods such as Taming 3DGS \cite{10.1145/3680528.3687694} are designed for static scenes, dynamic reconstruction additionally requires temporal and kinematic cues to preserve motion-critical primitives. To address this, we define a unified importance score $M_i$ that fuses geometric stability, kinematic significance, and perceptual impact:
\begin{equation}
\label{eq:importance}
M_i = \mathcal{N}\left(
\lambda_{gm} \cdot \mathcal{F}_{\text{geom/motion}}(g_i)
+   \mathcal{F}_{\text{perceptual}}(g_i)
\right),
\end{equation}
where $\mathcal{N}(\cdot)$ denotes Min-Max normalization that scales values to $[0,1]$, placing the heterogeneous cues on a common numerical scale before weighted aggregation. 

\textit{Geometric and Motion Cues} ($\mathcal{F}_{\text{geom/motion}}$).
This module captures the structural and kinematic necessity of a Gaussian. 
To handle the complexity of dynamic scenes, we explicitly decompose the score into five key components:

\begin{equation}
\begin{aligned}
\mathcal{F}_{\text{geom/motion}}(g_i) &= \mathbf{w}_1^T \cdot \mathcal{N}\bigg( \Big[ 
 \nabla^{\mu}_i, 
\alpha_i^{\text{max}},
  d_{i}^{-1}, \\
& \quad \quad \quad \quad \quad \quad  \quad\lambda_{\text{max}}(\Sigma_i), 
\mathrm{Mo}_i
\Big] \bigg),\\
 \nabla^{\mu}_i&=\left\|\nabla_{\boldsymbol{\mu}_i}\mathcal{L}_{\text{render}}\right\|_2,\\
\alpha_i^{\text{max}}&=\max_{1\leq t\leq T} \alpha_i(t),\\
 d_{i}^{-1} &= \frac{1}{|V|}\sum_{v} d_{i,v}^{-1},\\
 \mathrm{Mo}_i &= 
\begin{cases} 
\|\mathcal{T}_i\|_2, & \text{if } g_i \in \mathcal{S} \\
\|\mathbf{f}^{\mu}_i\|_2 + \|\mathbf{f}^R_i\|_2, & \text{if } g_i \in \mathcal{D}(t)
\end{cases}
\end{aligned}
\end{equation}

Here, $\alpha_i(t)$ denotes the opacity of Gaussian $i$ evaluated at timestamp $t$, including temporal activation when applicable.

For dynamic Gaussians, the norms of $\mathbf{f}^{\mu}_i$ and $\mathbf{f}^R_i$ are computed over the complete position and rotation trajectories.

The weighting vector $\mathbf{w}_1$ balances the contributions of the five cues, each of which captures a geometric or kinematic property relevant to high-fidelity reconstruction. Specifically, the Positional Gradient $\nabla^{\mu}_i$ quantifies the sensitivity of the reconstruction loss with respect to the Gaussian's position, effectively identifying primitives located in structure-critical regions where spatial precision is paramount. To capture temporal transients, Peak Opacity $\alpha_i^{\text{max}}$ utilizes the maximum opacity over the temporal sequence rather than the average, ensuring that fleeting structures are preserved rather than pruned due to low average visibility. We further incorporate Proximity ($d_{i}^{-1}$) via the inverse mean depth to explicitly prioritize foreground elements, as they occupy larger screen areas and demand higher rendering fidelity compared to distant background objects. To maintain background continuity, Spatial Extent $\lambda_{\text{max}}(\Sigma_i)$, derived from the largest eigenvalue of the covariance matrix, safeguards large-scale Gaussians that cover extensive regions, preventing the formation of geometric holes. Finally, the Motion Magnitude $\mathrm{Mo}_i$ employs a piece-wise formulation adapted to our decomposition: for static-branch Gaussians, it retains primitives needed to represent regular low-magnitude translation, while for dynamic Gaussians, it prioritizes those exhibiting large displacements and rotations to protect high-frequency motion details that are otherwise difficult to recover.

\textit{Perceptual Cues} ($\mathcal{F}_{\text{perceptual}}$).
To ensure visual fidelity, this component quantifies the rendering degradation caused by removing a Gaussian, aggregated across training views $V$:

\begin{equation}
\begin{aligned}
\mathcal{F}_\text{perceptual}(g_i) &= \mathbf{w}_2^T \cdot \mathcal{N}\bigg( \Big[
\mathcal{I}_i, \text{Area}_{i},
\text{Var}_i^{-1}
\Big] \bigg), \\
\mathcal{I}_i&=\sum_{v \in V} \|\mathcal{I}_v - \mathcal{I}_v^{\setminus \{i\}}\|_1, \\
\text{Area}_{i} &= \sum_{v \in V} \text{Area}_{i,v}, \\
\text{Var}_i^{-1} &=  \text{Var}_{v \in V}(\text{color}_{i,v})^{-1}.
\end{aligned}
\end{equation}

Here, $\mathbf{w}_2$ denotes the weighting vector that balances the contribution of each perceptual metric. Specifically, the Photometric Residual $\mathcal{I}_i$ directly quantifies the rendering degradation by aggregating the pixel-wise L1 error introduced to the rendered image when the Gaussian $g_i$ is excluded. To evaluate visual impact, Pixel Coverage $\text{Area}_{i}$ accumulates the projected screen area of the Gaussian across views, where larger footprints imply a greater contribution to the final appearance. Finally, Multi-view Consistency $\text{Var}^{-1}_i$ utilizes the inverse variance of the Gaussian's contribution across views to prioritize view-consistent primitives, effectively acting as a filter to suppress view-dependent noise or artifacts that would otherwise be visually distracting.

Both weighting vectors are fixed empirical hyperparameters rather than learnable parameters and are shared across all scenes and datasets without scene-specific tuning. Their numerical values are reported in Sec.~IV-A. Because the heterogeneous cues are normalized before aggregation, the coefficients specify the relative emphasis of the individual components.

\textbf{Densify-prune Policy Under Budget.}
Guided by the importance score $M_i$ and regulated by the current global target $N(r)$, our method performs densification and pruning in a unified framework. At each scheduled population update, eligible static and dynamic Gaussians are evaluated using the same score under this shared capacity; no manually predefined sub-budget is assigned to either branch. The controller then performs (i) \emph{densification}: Gaussians with high $M_i$ are cloned or split with higher probability, injecting detail where it is most useful; and (ii) \emph{pruning}: Gaussians with low $M_i$ are targeted for removal, gradually eliminating redundant or visually insignificant primitives.

This closed-loop controller dynamically directs the shared Gaussian capacity to the most informative
spatial and temporal regions while keeping the overall count near $N(r)$ and progressively approaching $N_{\text{target}}$.
Unlike heuristic grow-and-prune pipelines~\cite{lee2024ex4dgs,Wu_2024_CVPR}, 
our differentiable control embeds capacity regulation directly into optimization, 
enabling stable training and precise control over model complexity.

\subsection{Adaptive Static-Dynamic Allocation}
\label{sec:alloc}

Dynamic scenes are rarely uniformly active: while the majority of regions often remain static or undergo rigid, regular motion, significant non-rigid deformations are typically confined to specific objects within limited temporal spans. Consequently, uniformly distributing Gaussians across space and time results in a substantial waste of representational capacity. To fully utilize the fixed Gaussian budget $N_{\text{target}}$, we introduce an Adaptive Static-Dynamic Allocation strategy that determines and periodically updates the membership of the two representation branches defined in Sec.~\ref{sec:3.1}, without pre-labeled masks or a fixed static-dynamic ratio.

Our core insight is to identify a natural separation boundary in the distribution of Gaussian motion magnitudes. For each static-branch Gaussian, we measure its motion over the full trajectory as $m_i=\|\mathcal{T}_i\|_2$ and normalize it as $\widetilde{m}_i=\mathcal{N}(m_i)$. Camera ego-motion is accounted for by the calibrated camera poses and is therefore not interpreted as Gaussian-level scene motion. After the warm-up phase, we construct a motion histogram using $B$ bins and apply smoothing to suppress noise. Let $\mathbf{H}_{\mathcal{T}}$ denote the smoothed histogram and $\mathcal{P}_{\mathcal{T}}$ the set of candidate peaks:
\begin{equation}
\begin{aligned}
\mathbf{H}_{\mathcal{T}} &= \operatorname{smooth}(\operatorname{hist}(\{\widetilde{m}_i\},B)),\\
\mathcal{P}_{\mathcal{T}} &= \{p\mid \mathbf{H}_{\mathcal{T}}(p)>\mathbf{H}_{\mathcal{T}}(p\pm1)\}.
\end{aligned}
\end{equation}
Here, $p$ denotes the center of a histogram bin in the normalized motion domain.

When the histogram exhibits a stable two-population separation, we select the two most significant peaks and order them as $p_s<p_d$, corresponding to the lower- and higher-motion populations. Following this peak-valley criterion, the ordered peak pair and the intervening valley are determined by
\begin{equation}
\begin{aligned}
(p_s,p_d) &= \underset{p_1,p_2\in\mathcal{P}_{\mathcal{T}},\,p_1<p_2}{\operatorname{argmax}}
\left(\mathbf{H}_{\mathcal{T}}(p_1)+\mathbf{H}_{\mathcal{T}}(p_2)\right),\\
\tau_s &= \underset{x\in[p_s,p_d]}{\operatorname{argmin}}\,\mathbf{H}_{\mathcal{T}}(x).
\end{aligned}
\end{equation}
Here, $\tau_s$ denotes the initial separation in the histogram domain. To reduce sensitivity to histogram binning, we map this location back to the motion-data domain through its empirical percentile:
\begin{equation}
\alpha_{\mathcal{T}}=
\frac{\left|\{\widetilde{m}_i\mid\widetilde{m}_i<\tau_s\}\right|}
{\left|\{\widetilde{m}_i\}\right|},\qquad
\tau_{\text{motion}}=\operatorname{quantile}(\{\widetilde{m}_i\},\alpha_{\mathcal{T}}).
\end{equation}
For multi-modal distributions, this procedure seeks only the boundary between the lower-motion population and subsequent higher-motion populations rather than attempting to assign every motion mode separately. If no stable inter-peak valley can be identified, as in unimodal or weakly separable distributions, we instead use the fallback
\begin{equation}
\tau_{\text{motion}}=\operatorname{quantile}(\{\widetilde{m}_i\},q),
\end{equation}
where $q$ is automatically selected by a spread-adaptive rule: concentrated low-motion distributions use a higher quantile, whereas broader distributions use a lower one.

The resulting threshold is applied to visible Gaussians currently assigned to $\mathcal{S}$ during scheduled membership updates. A Gaussian is promoted to $\mathcal{D}(t)$ when its normalized motion exceeds $\tau_{\text{motion}}$, while dynamic-to-static conversion is not performed. For a promoted Gaussian, the dynamic trajectory is initialized from the temporal positions generated by its original center and translational attribute at the corresponding timestamps. Its SH features, scale, opacity, and rotation are inherited, with the rotation replicated across the temporal representation, and its temporal activation parameters are initialized using the corresponding timestamp information. The original static-branch Gaussian is then removed, making the migration a one-to-one replacement that changes only branch membership without altering the total Gaussian count. Subsequent densification and pruning of both branches remain governed by the shared global controller introduced in Sec.~\ref{sec:budget}.

\subsection{Training Strategy}
\label{sec:training}

Training our constrained dynamic Gaussian model involves jointly optimizing scene appearance, differentiable budget control, and adaptive allocation in a stable and progressive manner. We adopt a three-phase training strategy designed to (i) obtain a reliable initialization, (ii) introduce differentiable population control, and (iii) stabilize optimization while enforcing the capacity constraint.

\textbf{Phase I: Warm-up and Initialization.}
We begin with a short warm-up stage that initializes the representation without budget constraints.
We utilize static-branch Gaussians equipped with lightweight per-Gaussian translational attributes, allowing the model to establish scene geometry and regular-motion priors using only the rendering loss $\mathcal{L}_{\text{render}}$ (Eq.~\ref{rendering_loss}).
These pre-trained Gaussians serve as a foundational prior, providing reliable geometric and kinematic guidance for the subsequent modules.

\textbf{Phase II: Differentiable Budget Enforcement.}
After warm-up, we activate the population controller (Sec.~\ref{sec:budget}) and introduce the budget loss $\mathcal{L}_{\text{budget}}^{(r)}$ and regularization $\mathcal{L}_{\text{reg}}$.
The total loss is formulated as:
\begin{equation}
\mathcal{L} = \mathcal{L}_{\text{render}} + \lambda_b \mathcal{L}_{\text{budget}}^{(r)} + \lambda_r \mathcal{L}_{\text{reg}}
\label{eq:total_loss}
\end{equation}
where $\lambda_b$ and $\lambda_r$ control the strength of the budget penalty and regularization, respectively. The regularization term $\mathcal{L}_{\text{reg}}$ jointly encourages temporal smoothness and compact Gaussian-attribute distributions, thereby benefiting both dynamic reconstruction and subsequent compression.
Crucially, the Adaptive Static-Dynamic Allocation (Sec.~\ref{sec:alloc}) is performed periodically alongside densification, pruning, and the update of the global sub-target $N(r)$.
At each interval, static-to-dynamic membership is updated and both branches compete under the same global capacity rather than receiving predefined sub-budgets.
Simultaneously, discrete pruning is triggered for Gaussians with low opacity or near-zero mask values ($c_i \approx 0$), while the gate temperature is annealed to gradually enforce a binary selection.

\textbf{Phase III: Stabilization and Fine-tuning.}
After the scheduled budget-enforcement phase brings the effective count $N_p$ close to $N_{\text{target}}$, we enter the stabilization phase.
The soft masks $c_i$ are explicitly binarized, and redundant Gaussians (where $c_i=0$) are permanently culled to strictly satisfy the budget constraint.
The remaining compact set is then fine-tuned solely with $\mathcal{L}_{\text{render}}$ and $\mathcal{L}_{\text{reg}}$ to recover any detail loss caused by the masking process and to ensure temporal smoothness.

\subsection{Dual-Mode Hybrid Compression}
\label{sec:compress}

To further reduce storage and transmission cost under the fixed Gaussian budget,
we adopt a dual-mode hybrid compression strategy tailored to the static and dynamic components
introduced in Sec.~\ref{sec:alloc}.
\begin{figure}[t]
\centering
\includegraphics[width=\linewidth]{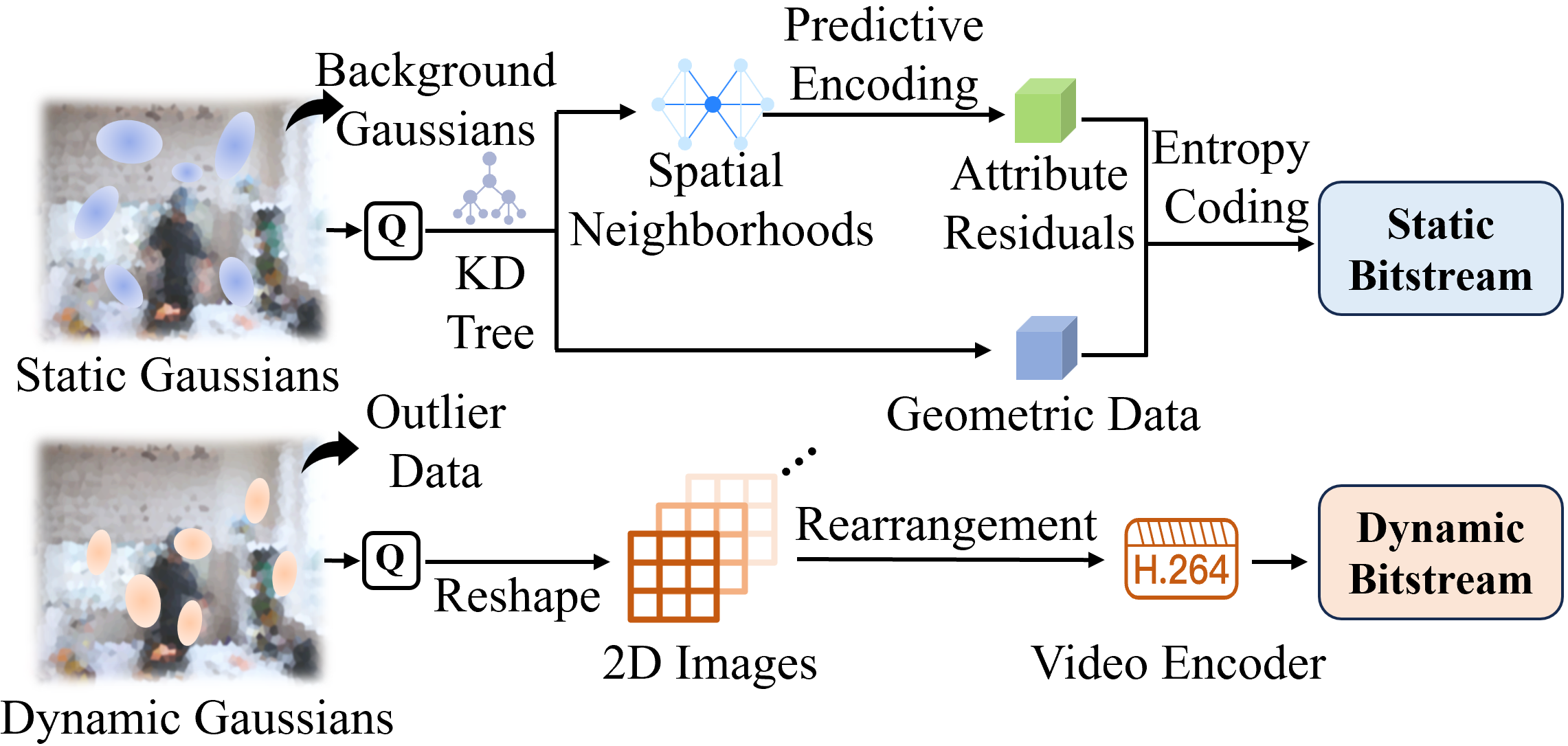}

\caption{Illustration of our dual-mode hybrid compression strategy, which separates outliers from both static and dynamic Gaussian data, then applies distinct compression approaches.}
\label{fig:Compress}

\end{figure}

\textbf{Static Compression.} 
For static Gaussians, we observe that outliers far from the scene center significantly degrade the post-compression quality. This is because these distant points excessively expand the coordinate bounding box. Under uniform quantization, an expanded range forces a larger quantization step size, which drastically reduces the precision allocated to the densely populated foreground regions. To mitigate this, we first compute the mean $\mu_d$ and standard deviation $\sigma_d$ of all Gaussian distances to the scene center and separate background Gaussians using a threshold of $\mu_d + 3\sigma_d$. For the remaining foreground, after attribute quantization, we employ a KD-tree to spatially reorder points, ensuring local continuity. Leveraging these spatial neighborhoods, we apply predictive encoding to residuals to reduce entropy. Finally, geometric data and attribute residuals undergo entropy encoding to complete the compression.

\textbf{Dynamic Compression.} 
For dynamic Gaussians, we first conduct a statistical analysis of the data distribution across channels, observing that: (1) distributions within the same attribute are highly similar across channels; and (2) each attribute is generally concentrated but contains significant outliers. Based on these findings, we design a dedicated compression pipeline. First, we separate top 5\% of outlier data to narrow the value range and improve concentration. Next, we reshape each quantized attribute of the dynamic Gaussians into a 2D image format and rearrange them by grouping identical attributes together. These organized video sequences are then encoded using an H.264 encoder (based on the x264 library). To ensure high fidelity and temporal consistency, the encoder is configured with the YUV 4:4:4 color space, the “medium” preset, and a constant Quantization Parameter of 20. Furthermore, we restrict the encoding to I- and P-frames (disabling B-frames) with 3 reference frames to optimize for decoding efficiency.

\section{Experiments}
\subsection{Configurations}
\textbf{Datasets.}
We evaluate our method on three challenging real-world datasets representing different dynamic characteristics. 
First, the N3DV dataset \cite{li2022neural} provides 6 scenes with complex non-rigid deformations, captured by 18-21 cameras at $2704\times2028$ resolution and 30 FPS. 
Second, the Technicolor dataset \cite{attal2023hyperreel} offers studio-quality scenes recorded by a 16-camera rig at $2048\times1088$, featuring rich textures and lighting effects. 
Third, the MeetRoom dataset \cite{streaming} captures long-duration human activities in an indoor setting using 14 cameras at $1280\times720$ and 30 FPS. 
Consistent with prior works \cite{sun20243dgstream,lee2024ex4dgs}, we employ a standard leave-one-out strategy. For the N3DV and Meetroom datasets, the first camera view is held out for testing. In contrast, for the Technicolor dataset, we reserve the camera positioned at the intersection of the second row and second column for evaluation. All remaining views serve as the training set.

\textbf{Implementation.}
The experiments were conducted on hardware with an Intel(R) Xeon(R) W-2245 CPU @ 3.90 GHz and an RTX 3090 graphics card. For each sequence, our three-phase training strategy consists of 500, 29500, and 10000 iterations, respectively, with Gaussian densification, pruning, membership updates, and global sub-target updates performed every 500 steps during the second phase. $\lambda_{\text{ssim}}=0.2$, $\lambda_b = 1\times10^{-7}$, $\lambda_{r} = 1\times10^{-4}$ and $\lambda_{gm}=2$ were set for all sequences. We use the fixed weighting vectors $\mathbf{w}_1=[5,20,1,5,10]$ and $\mathbf{w}_2=[10,2,10]$ for all scenes and datasets without scene-specific tuning. The gate temperature is annealed from $\tau_{\text{init}}=1.0$ to $\tau_{\text{end}}=0.01$, and the quantile fallback for adaptive static-dynamic separation automatically selects $q$ within $[0.8,0.99]$. In the compression phase, 16-bit quantization was used for $\boldsymbol{\mu}$, $\mathcal{T}$ and $\mathbf{f}_{\mu}$, while 8-bit quantization was applied to all other attributes.

\textbf{Metrics.} To assess the modeling and control capabilities of our method across experimental datasets, we adopt Peak Signal-to-Noise Ratio (PSNR) and Structural Similarity Index (SSIM) \cite{1284395} as quality metrics, along with model size measured in megabytes for the entire sequence. For comprehensive rate-distortion performance analysis, we apply Bjontegaard Delta PSNR (BD-PSNR) \cite{2007An}. Rendering efficiency is gauged by frames per second (FPS). To validate our method's capacity for Gaussian count control, we further report the number of Gaussians and the error ratio relative to the target.

\subsection{Comparison}
\begin{table}[t]
\centering
\setlength{\tabcolsep}{2.5pt} 
\renewcommand{\arraystretch}{1.08}
\caption{Quantitative comparison on the N3DV \cite{li2022neural} dataset. The PSNR, SSIM, and rendering speed are averaged over all 300 frames for each scene. The reported model size is the total for the entire sequence. Ours-l and Ours-s use target Gaussian numbers of 300,000 and 100,000, respectively. \textbf{Best} and \underline{second-best} results are highlighted.}

\label{t1}
\resizebox{\columnwidth}{!}{
\begin{tabular}{c|cccc|c}
\toprule[1pt]
Method      & \begin{tabular}[c]{@{}c@{}}PSNR$\uparrow$\\ (dB)\end{tabular} & SSIM$\uparrow$  & \begin{tabular}[c]{@{}c@{}}Size$\downarrow$\\ (MB)\end{tabular} & \begin{tabular}[c]{@{}c@{}}Render$\uparrow$\\ (FPS)\end{tabular} & \makecell{User-Defined\\Gaussian-Count\\Control} \\ \hline
K-Planes\cite{fridovich2023k}    & 29.91    & 0.920 & 300      & 0.15        & \ding{53}           \\
ReRF\cite{rerf}        & 29.71    & 0.918 & 231      & 2.0         & \ding{53}           \\
TeTriRF\cite{tetrirf}     & 30.65    & 0.931 & 227      & 2.7         & \ding{53}           \\
StreamRF\cite{streaming}    & 30.61    & 0.930 & 2280     & 8.3         & \ding{53}           \\
3DGStream\cite{sun20243dgstream}   & 31.54    & 0.942 & 2430     & \textbf{215}         & \ding{53}           \\
4DGC\cite{hu20254dgc}        & 31.58    & 0.943 & 150      & 168         & \ding{53}           \\
4DGCPro\cite{zheng20254dgcpro}        & 31.64    & 0.943 & 192      & 170         & \ding{53}           \\
4DGaussians\cite{Wu_2024_CVPR} & 31.15    & 0.939 & 34       & 147         & \ding{53}           \\
4DGS\cite{yang2023gs4d}        & 32.01    & 0.944 & 6270     & 72          & \ding{53}           \\
STGS\cite{Li_STG_2024_CVPR}        & 32.05    & 0.944 & 200      & 107         & \ding{53}           \\
RD4DGS\cite{lee2025rd4dgs}  & 29.66    & 0.917 & 11.1      & 100.9         & \ding{53} \\
Swift4D\cite{wu2025swift4d}  & 31.79    & 0.944 & 30      & 128         & \ding{53} \\
GIFStream\cite{li2025gifstream4dgaussianbasedimmersive}  & 31.75    & 0.938 & \underline{10}      & 95         & \ding{53} \\
ReCon-GS\cite{fu2025neurips-recongs} & \underline{32.12} & \underline{0.945} & 62 & 166 & \ding{53} \\
4D Scaffold GS\cite{cho20254dscaffoldgaussiansplatting} & 32.03 & \underline{0.945} & 149 & 130 & \ding{53} \\
StreamSTGS\cite{ke2025edgegaussian} & 31.68 & 0.943 & 87.95 & 100 & \ding{53} \\
Ex4DGS\cite{lee2024ex4dgs}      & 32.11    & \underline{0.945} & 115      & 128         & \ding{53}           \\ \hline
Ours-l      & \textbf{32.14}    & \textbf{0.946} & 31.5     & 149         & \checkmark          \\
Ours-s      & 31.83    & 0.944 & \textbf{6.8}      & \underline{186}         & \checkmark         \\ \bottomrule[1pt]
\end{tabular}
}

\end{table}

\begin{table}[t]
\centering
\setlength{\tabcolsep}{5pt} 
\renewcommand{\arraystretch}{1.15}
\caption{Validation of our precise control capability over the total number of Gaussians. We report the average PSNR on the coffee\_martini sequence, the number of static/dynamic/overall Gaussians, and the error ratio of Gaussian number.}

\label{t4}
\scalebox{1}{
\begin{tabular}{c|c|ccccc}
\toprule[1pt]
Method                & Target & PSNR  & Static & Dynamic & Overall & Ratio   \\ \hline
Ex4DGS\cite{lee2024ex4dgs}                & -      & 28.79 & 292.2k & 47.7k   & 339.9k  & -      \\ \hline
\multirow{4}{*}{Ours} & 100k   & 28.53 & 72.4k  & 27.6k   & 99.9k   & 0.1\% \\
                      & 200k   & 28.68 & 156.8k & 41.0k   & 197.8k  & 1.1\% \\
                      & 300k   & 28.81 & 244.4k & 51.4k   & 295.8k  & 1.4\% \\
                      & 400k   & 28.95 & 316.4k & 81.2k   & 397.6k  & 0.6\% \\
                      \bottomrule[1pt]
\end{tabular}
}

\end{table}
\textbf{Quantitative Comparisons.} To validate the effectiveness of our method, we compare it against a broad range of state-of-the-art dynamic reconstruction and compression approaches, particularly 3DGS-based methods, including 3DGStream \cite{sun20243dgstream}, 4DGC \cite{hu20254dgc}, 4DGCPro \cite{zheng20254dgcpro}, 4DGaussians \cite{Wu_2024_CVPR}, 4DGS \cite{yang2023gs4d}, STGS \cite{Li_STG_2024_CVPR}, RD4DGS \cite{lee2025rd4dgs}, Swift4D \cite{wu2025swift4d}, GIFStream \cite{li2025gifstream4dgaussianbasedimmersive}, Ex4DGS \cite{lee2024ex4dgs}, ReCon-GS \cite{fu2025neurips-recongs}, 4D Scaffold Gaussian Splatting \cite{cho20254dscaffoldgaussiansplatting}, and StreamSTGS \cite{ke2025edgegaussian}. We present two variants, \textbf{Ours-l} and \textbf{Ours-s}, obtained by setting different target Gaussian numbers to demonstrate scalability. As summarized in Tab. \ref{t1}, Ours-l achieves the highest PSNR and SSIM while requiring only \textbf{31.5 MB}. Methods such as 4DGS, STGS, and Ex4DGS achieve comparable reconstruction quality but require substantially larger representations, whereas compact approaches such as 4DGaussians, RD4DGS, Swift4D, and GIFStream incur varying degrees of quality degradation. ReCon-GS also obtains comparable reconstruction quality but requires nearly twice the storage of Ours-l, while 4D Scaffold GS and StreamSTGS require 149 MB and 87.95 MB, respectively. Moreover, CDGS is the only method in the comparison that directly supports a user-defined Gaussian-count target during optimization.
More importantly, we can achieve precise control over the total number of Gaussians. As shown in Tab. \ref{t4}, under different target total numbers of Gaussians, we maintain the error rate within \textbf{2\%} while adaptively determining the static-dynamic composition.

\begin{table}[]
\centering
\setlength{\tabcolsep}{4pt} 
\renewcommand{\arraystretch}{1.15}
\caption{Quantitative comparison on the MeetRoom dataset \cite{streaming} and Technicolor dataset \cite{attal2023hyperreel}.}

\label{t2}
\scalebox{1}{
\begin{tabular}{c|ccccc}
\toprule[1pt]
Dataset                                                                     & Method     & \begin{tabular}[c]{@{}c@{}}PSNR$\uparrow$\\ (dB)\end{tabular} & SSIM$\uparrow$  & \begin{tabular}[c]{@{}c@{}}Size$\downarrow$\\ (MB)\end{tabular} & \begin{tabular}[c]{@{}c@{}}Render$\uparrow$\\ (FPS)\end{tabular}  \\ \hline
\multirow{10}{*}{\begin{tabular}[c]{@{}c@{}}MeetRoom\\ Dataset\cite{streaming}\end{tabular}} & ReRF\cite{rerf}       & 26.43    & 0.911 & 189      & 2.9         \\
                                                                            & TeTriRF\cite{tetrirf}    & 27.37    & 0.917 & 183      & 3.8         \\
                                                                            & StreamRF\cite{streaming}   & 26.71    & 0.913 & 2469     & 10          \\
                                                                            & 3DGStream\cite{sun20243dgstream}  & 28.03    & 0.921 & 2430     & \textbf{288}         \\
                                                                            & 4DGC\cite{hu20254dgc}       & 28.08    & 0.922 & 126      & 213         \\
                                                                            & 4DGCPro\cite{zheng20254dgcpro}       & 28.02    & 0.921 & 123      & \underline{222}         \\
                                                                            & STGS\cite{Li_STG_2024_CVPR}  &  29.01    &   0.929      & 15.2       &  159                     \\
                                                                            & Ex4DGS\cite{lee2024ex4dgs}     & \underline{29.12}    & \underline{0.930} & 40.2     & 148         \\ \cline{2-6}
                                                                            & Ours-l     & \textbf{29.18}    & \textbf{0.931} & \underline{10.4}     & 165         \\
                                                                            & Ours-s     & 28.60    & 0.927 & \textbf{4.5}      & 215         \\ \hline
\multirow{7}{*}{Technicolor\cite{attal2023hyperreel}}                                                & 4DGC\cite{hu20254dgc}        & 31.56    & 0.939      &   310       &   \underline{139}          \\
& 4DGCPro\cite{zheng20254dgcpro}        & 31.53    & 0.939      &   453       &   133          \\
& STGS\cite{Li_STG_2024_CVPR}        & 31.96    & \underline{0.941}      &   58.7       &   113          \\
& RD4DGS\cite{lee2025rd4dgs}     & 31.33    &  0.938     &    \underline{33.6}      &  123           \\ 
                                                                            & Ex4DGS\cite{lee2024ex4dgs}     & \underline{32.38}    &  \textbf{0.942}     &    170.3      &  100           \\ \cline{2-6}
                                                                            & Ours-l     &   \textbf{32.41}       &  \textbf{0.942}     &  48.0        &    125         \\ 
                                                                            & Ours-s     &   31.90       &  0.940     &  \textbf{17.1}        &    \textbf{144}         \\ \bottomrule[1pt]
\end{tabular}
}
\end{table}

\begin{figure*}[t]
\centering
\includegraphics[width=\linewidth]{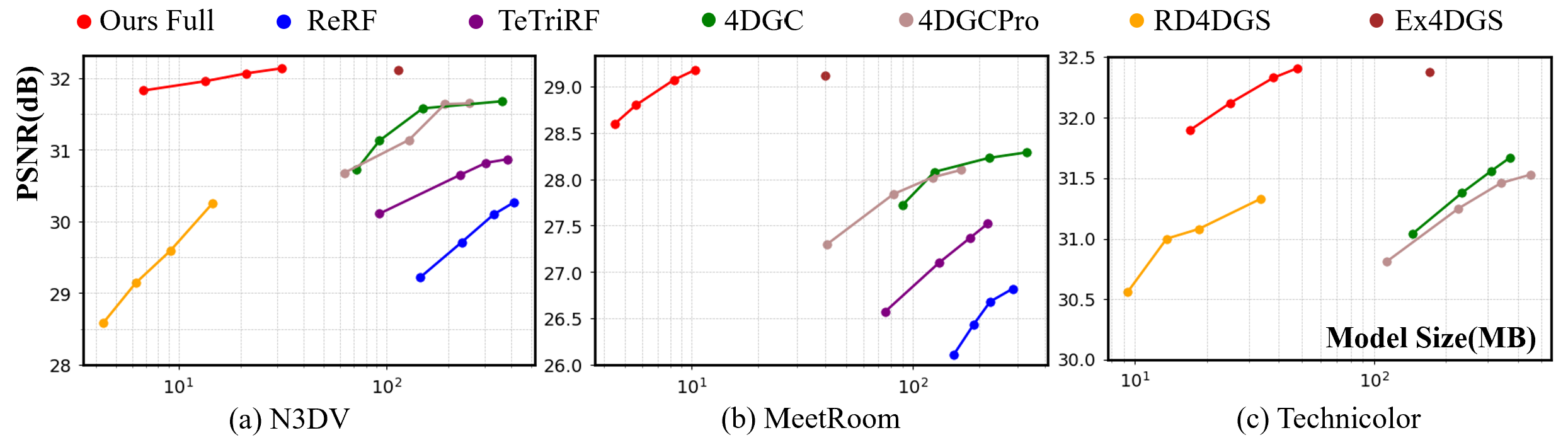}
\caption{Rate-distortion curves on different datasets, illustrating the superiority of our method over ReRF \cite{rerf}, TeTriRF \cite{tetrirf}, 4DGC \cite{hu20254dgc}, 4DGCPro \cite{zheng20254dgcpro}, RD4DGS \cite{lee2025rd4dgs}, and Ex4DGS \cite{lee2024ex4dgs}.}
\label{fig:rd}
\end{figure*}

We further validate the generality of our method on the MeetRoom \cite{streaming} and Technicolor  \cite{attal2023hyperreel} datasets. As summarized in Tab. \ref{t2}, our approach consistently outperforms a wide range of methods in PSNR, SSIM, and model size, corroborating its robustness across diverse scenes.
We next evaluate the rate-distortion performance against contemporary works, including ReRF \cite{rerf}, TeTriRF \cite{tetrirf}, 4DGC \cite{hu20254dgc}, 4DGCPro \cite{zheng20254dgcpro}, and RD4DGS \cite{lee2025rd4dgs}. As visualized in the RD curves of Fig. \ref{fig:rd}, our approach consistently pushes the Pareto frontier, outperforming variable-bitrate baselines across the spectrum. This superiority is also quantitatively corroborated by the BD-PSNR results in Tab. \ref{t3}. Our method demonstrates the highest gains, with improvements over 4DGC of \textbf{1.90 dB} and \textbf{1.72 dB} on the N3DV and MeetRoom datasets, which is a substantial margin above all other comparisons.

\begin{table}[]
\centering
\setlength{\tabcolsep}{1pt} 
\renewcommand{\arraystretch}{1.15}
\caption{The BD-PSNR(dB) results of our CDGS, ReRF \cite{rerf}, TeTriRF \cite{tetrirf}, 4DGCPro \cite{zheng20254dgcpro} and RD4DGS \cite{lee2025rd4dgs} when compared with 4DGC \cite{hu20254dgc} on different datasets.}

\label{t3}
\scalebox{0.96}{
\begin{tabular}{c|cccc|c}
\toprule[1pt]
Dataset  & ReRF\cite{rerf}  & TeTriRF\cite{tetrirf} & 4DGCPro\cite{zheng20254dgcpro} & RD4DGS\cite{lee2025rd4dgs} & Ours \\ \hline
N3DV\cite{li2022neural}     & -1.99 & -1.12   & 0.08  & 1.33  & \textbf{1.90} \\
MeetRoom\cite{streaming} & -1.84 & -0.86   & -0.02      & - & \textbf{1.72} \\ \bottomrule[1pt]
\end{tabular}
}

\end{table}

Tab. \ref{t5} presents a comparison of computational efficiency between our CDGS and several dynamic scene reconstruction and compression methods. Our CDGS exhibits significantly improved computational efficiency: its training time is \textbf{1.0 hour}, compared to \textbf{1.2 hours} for Ex4DGS \cite{lee2024ex4dgs}, owing to the increased interval of Gaussian expansion. Additionally, experimental results on the N3DV dataset demonstrate that our approach enables high-speed rendering, achieving a rate of \textbf{5.4 ms} per frame. For encoding and decoding, CDGS achieves times of \textbf{16 s} and \textbf{0.5 s} for 300 frames in total, respectively, outperforming all other methods by a significant margin. These results fully demonstrate that CDGS exhibits comprehensively fast performance across reconstruction, rendering, and encoding/decoding.

\begin{table}[]
\centering
\setlength{\tabcolsep}{3pt} 
\renewcommand{\arraystretch}{1.15}
\caption{Complexity comparison of our method with dynamic scene reconstruction and compression methods.}

\label{t5}
\resizebox{\columnwidth}{!}{
\begin{tabular}{c|cccc|c}
\toprule[1pt]
Time      & ReRF\cite{rerf}              & TeTriRF\cite{tetrirf} & 4DGC\cite{hu20254dgc}  &Ex4DGS\cite{lee2024ex4dgs} & Ours \\ \hline
Encode(s) & 246               & 219     & 810      & -     & \textbf{16}   \\
Decode(s) & 18.3              & 16.8    & 28.2      & -     & \textbf{0.5}  \\
Train(h)  & \textgreater{}100 & 5.2     & 4.2     & 1.2   & \textbf{1.0}  \\
Render(ms)  & 497 & 372    & 5.6     & 7.8   & \textbf{5.4}  \\\bottomrule[1pt]
\end{tabular}
}

\end{table}

\begin{figure*}[t]
\centering
\includegraphics[width=\linewidth]{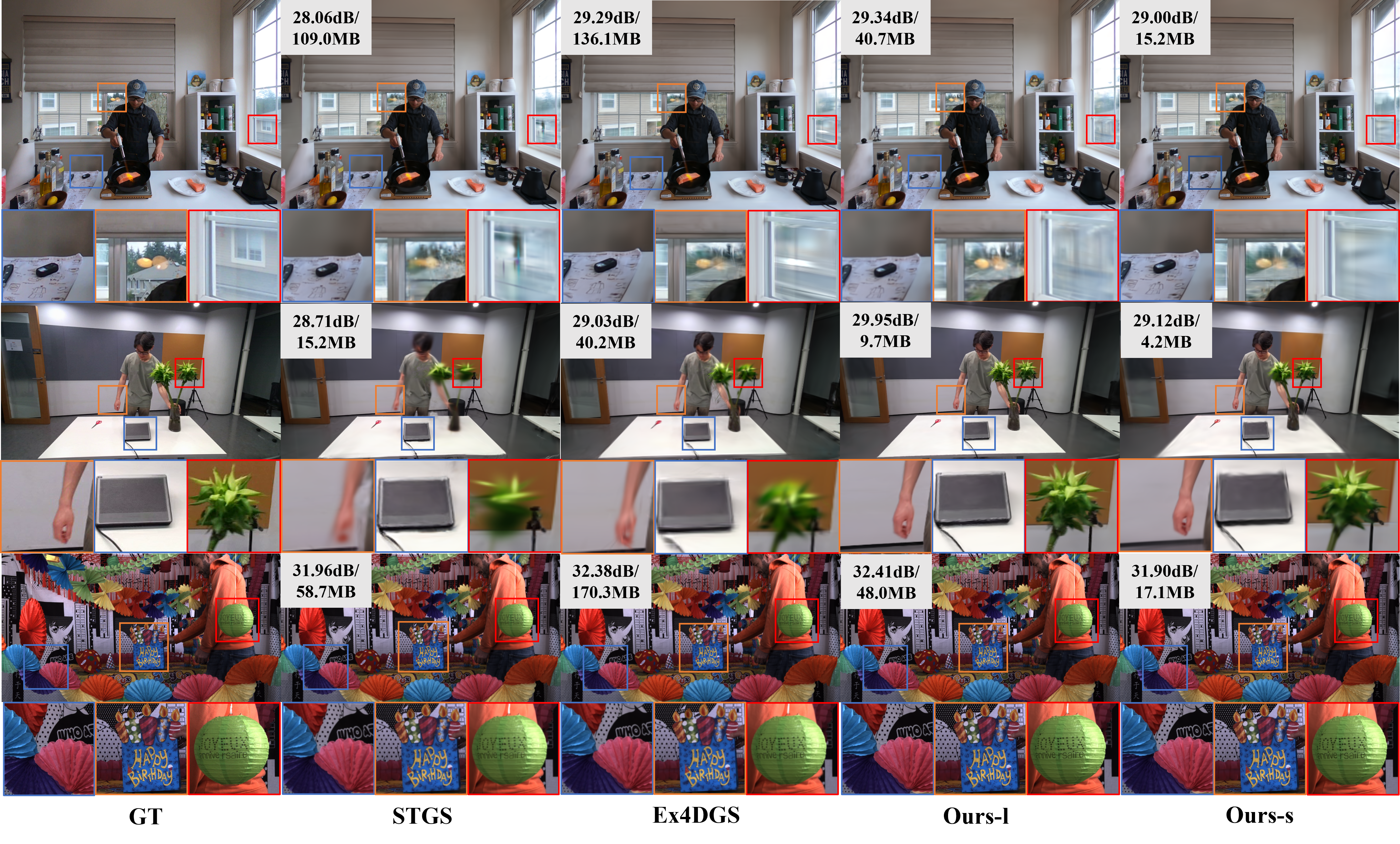}

\caption{Qualitative comparison of our CDGS against STGS \cite{Li_STG_2024_CVPR} and Ex4DGS \cite{lee2024ex4dgs} on the N3DV \cite{li2022neural}, MeetRoom \cite{streaming} and Technicolor \cite{attal2023hyperreel} datasets, demonstrating the performance of CDGS under different target Gaussian numbers. 
}
\label{fig:compare}

\end{figure*}

\textbf{Qualitative Comparisons.} As shown in Fig. \ref{fig:compare}, we qualitatively compare our method at different Gaussian number targets against STGS \cite{Li_STG_2024_CVPR} and Ex4DGS \cite{lee2024ex4dgs} on the \textit{flame\_salmon} sequence, the \textit{trimming} sequence, and the \textit{birthday} sequence. Our method delivers reconstruction quality comparable to these approaches while operating at a significantly lower model size. Compared to STGS, CDGS better preserves fine-grained details, such as the head, window, and desk in \textit{flame\_salmon}, the fast-moving hands and intricate plants in \textit{trimming}, as well as balloons, fans, and candles in \textit{birthday}. Furthermore, our approach enables precise Gaussian count control, consistently yielding optimal performance across varying total Gaussian numbers. For instance, in our lightweight variant (ours-s), reconstructing the \textit{flame\_salmon} sequence requires only \textbf{10k} Gaussians, which is far fewer than the \textbf{35k} used by Ex4DGS. This underscores that CDGS not only accurately captures dynamic scene content and retains high-fidelity details in complex objects but also achieves a highly compact model size, all while realizing precise regulation of the Gaussian count.

\subsection{Evaluations}

\begin{figure}[t]
\centering
\includegraphics[width=\linewidth]{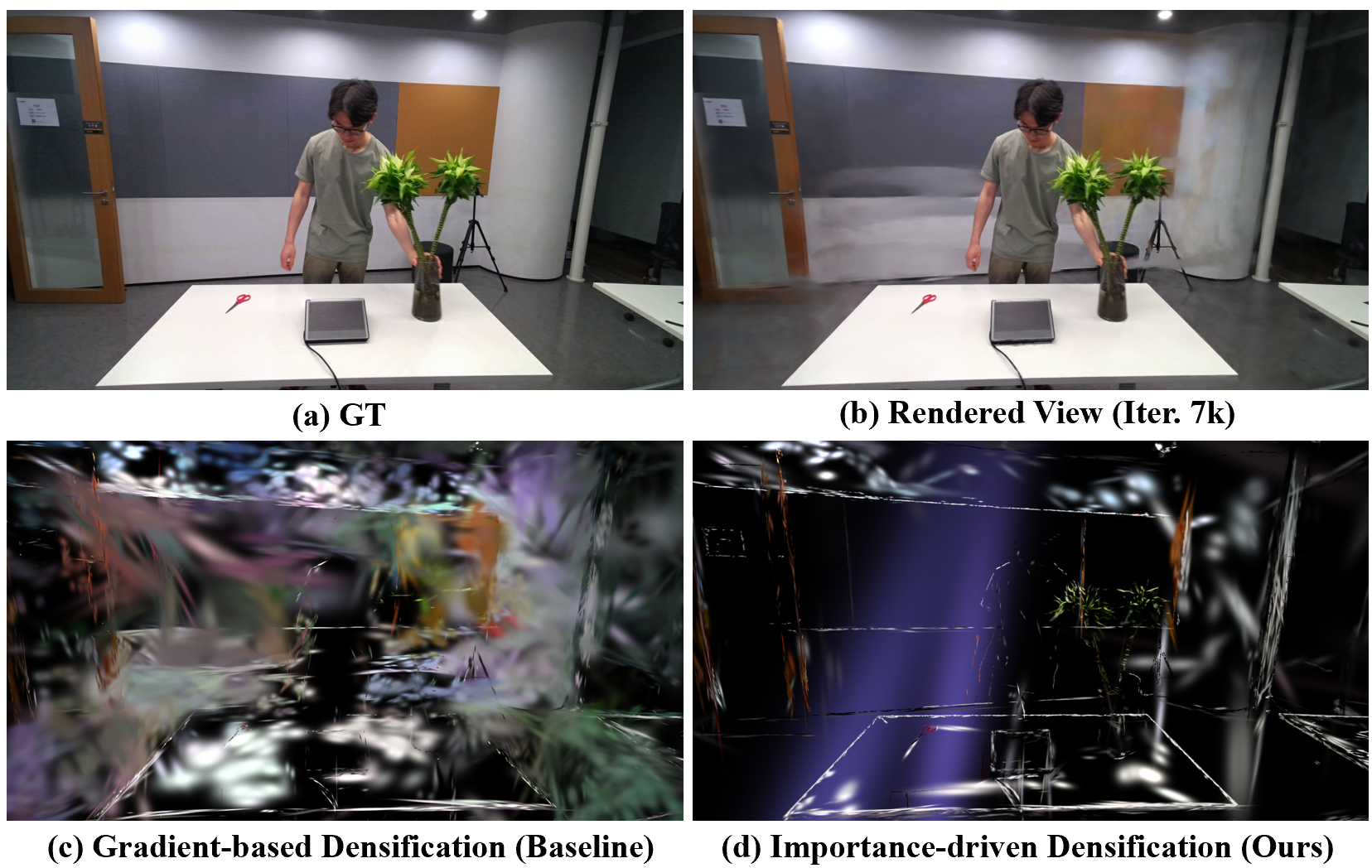}

\caption{Visualization of densification strategies at Iter. 7k. Unlike the gradient-based baseline \textbf{(c)} which generates excessive redundant primitives in the background, our importance-driven method \textbf{(d)} effectively suppresses irrelevant growth, concentrating Gaussians strictly on geometric and motion boundaries.}
\label{fig:IS}

\end{figure}

\begin{table}[t]
\centering
\setlength{\tabcolsep}{8pt} 
\renewcommand{\arraystretch}{1.15}
\caption{Ablation study of differentiable budget control and adaptive static-dynamic allocation.}

\label{t6}
\scalebox{1}{
\begin{tabular}{c|ccc}
\toprule[1pt]
                        & PSNR(dB)$\uparrow$ & Size(MB)$\downarrow$ & Ratio$\downarrow$ \\ \hline
w/o Importance score    &  31.91    &  31.2    & 1.6\%      \\
w/o $\mathcal{F}_{\text{geom/motion}}$ &  31.97    &  31.5    & 1.3\%      \\
w/o $\mathcal{F}_{\text{perceptual}}$  &  31.99    &  31.3    & 1.4\%      \\
$\lambda_{gm}=1.5$  &  32.08    &  31.4    & 1.3\%     \\
$\lambda_{gm}=2.5$  &  32.11    &  31.5    & 1.4\%      \\
w/o Budget loss         &  32.15    &  31.7    & 4.8\%      \\
w/o Adaptive allocation &  31.96    &  33.2    & 1.3\%      \\
Ours full               &  32.14    &  31.5    & 1.3\%      \\ \bottomrule[1pt]
\end{tabular}
}

\end{table}

\begin{table}[t]
\centering
\setlength{\tabcolsep}{3pt}
\renewcommand{\arraystretch}{1.12}
\caption{Component-wise ablation of the Unified Importance Score on N3DV \cite{li2022neural} and MeetRoom \cite{streaming}. PSNR and model size are averaged over all scenes.}
\label{t6_component}
\begin{tabular}{l|cc|cc}
\toprule[1pt]
& \multicolumn{2}{c|}{N3DV \cite{li2022neural}} & \multicolumn{2}{c}{MeetRoom \cite{streaming}} \\
& PSNR(dB) & Size(MB) & PSNR(dB) & Size(MB) \\ \hline
w/o $\nabla_i^{\mu}$ & 32.05 & 31.8 & 29.08 & 10.6 \\
w/o $\alpha_i^{\max}$ & 32.07 & 31.5 & 29.07 & 10.3 \\
w/o $d_i^{-1}$ & 32.09 & 31.5 & 29.12 & 10.7 \\
w/o $\lambda_{\max}(\Sigma_i)$ & 32.08 & 31.6 & 29.10 & 10.4 \\
w/o $\mathrm{Mo}_i$ & 32.02 & 31.3 & 29.02 & 10.5 \\ \hline
w/o $\mathcal{I}_i$ & 32.03 & 31.7 & 29.03 & 10.4 \\
w/o $\mathrm{Area}_i$ & 32.08 & 31.5 & 29.10 & 10.4 \\
w/o $\mathrm{Var}_i^{-1}$ & 32.06 & 31.6 & 29.07 & 10.3 \\ \hline
Ours Full & 32.14 & 31.5 & 29.18 & 10.4 \\ \bottomrule[1pt]
\end{tabular}

\end{table}

\begin{figure*}[t]
\centering
\includegraphics[width=\linewidth]{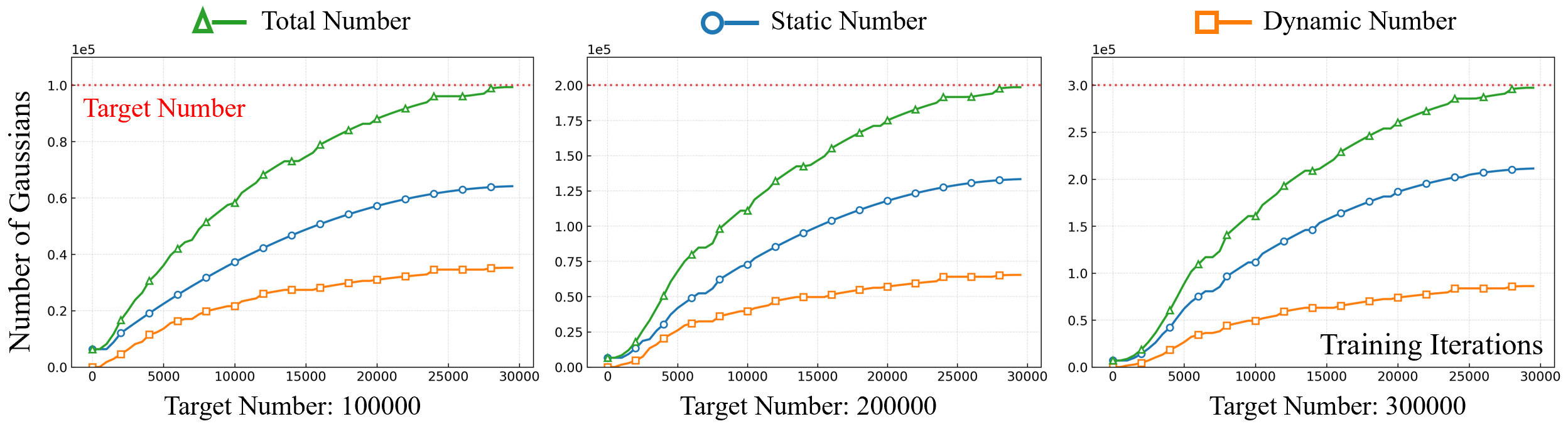}
\caption{Evolution of Gaussian counts under varying target budgets. The subplots correspond to target limits of $1 \times 10^5$ (\textbf{Left}), $2 \times 10^5$ (\textbf{Middle}), and $3 \times 10^5$ (\textbf{Right}). The curves illustrate the growth trajectories of \textcolor{mplblue}{\textbf{Static}}, \textcolor{mplorange}{\textbf{Dynamic}}, and \textcolor{mplgreen}{\textbf{Total}} Gaussians. Under our constrained optimization framework, the total number steadily increases and converges precisely to the predefined \textcolor{mplred}{Target Number}.}
\label{fig:growth_curve}
\end{figure*}

\textbf{Gaussian Number Constraint.} 
We conduct an ablation study to quantify the contribution of our proposed importance score and budget loss to Gaussian count control. As presented in Tab. \ref{t6}, the unified importance score, which evaluates each Gaussian's actual rendering contribution, is more effective than relying on traditional training gradients. 
This superiority is visually explicated in Fig. \ref{fig:IS}. As shown in Fig. \ref{fig:IS} (c), the gradient-based baseline struggles to distinguish between structural details and background noise, leading to blind densification in empty regions. Conversely, our method (Fig. \ref{fig:IS} (d)) acts as a precise filter, concentrating the Gaussian budget solely on perceptually salient object boundaries.
Ablating the entire importance score causes a significant performance drop of \textbf{0.23 dB}. The individual components of the score are also validated: removing the geometric/motion term ($\mathcal{F}_{\text{geom/motion}}$) leads to a \textbf{0.17 dB} loss, while omitting the perceptual term ($\mathcal{F}_{\text{perceptual}}$) results in a \textbf{0.15 dB} degradation. We further conduct a sensitivity analysis on the weighting coefficient $\lambda_{gm}$. Our experiments indicate that deviating from the default setting leads to varying degrees of quality degradation. This confirms that a balanced trade-off between geometric/motion stability and perceptual fidelity is essential for maximizing the utility of the limited Gaussian budget. Furthermore, the budget loss is proven critical for regulating the score distribution during training, ensuring the Gaussian count aligns with the target. Without it, the Gaussian count error increases by \textbf{3.5\%}. This precise controllability is visually corroborated in Fig. \ref{fig:growth_curve}. As observed, the total number of Gaussians monotonically increases and converges exactly to the preset limits without overshooting, demonstrating the stability and strictness of our budget enforcement strategy.

The component-level results in Tab.~\ref{t6_component} further show that removing any geometric/motion or perceptual cue reduces PSNR on both datasets under the same target count and compression pipeline. Motion magnitude and photometric residual have the largest individual effects, supporting the complementary contributions of both groups to the unified importance score.

\begin{figure}[]
\centering
\includegraphics[width=\linewidth]{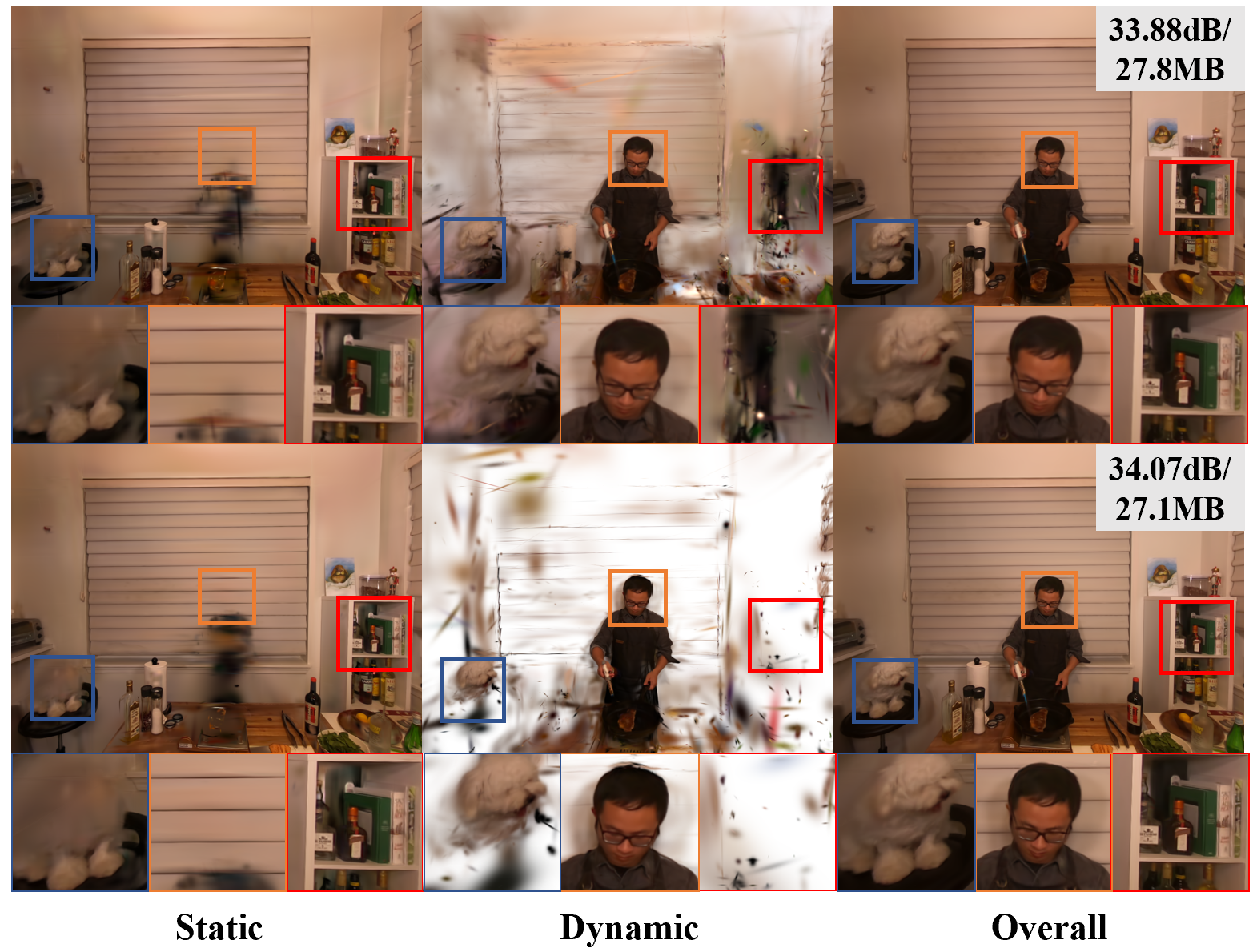}

\caption{Visualization of our adaptive static-dynamic allocation strategy (\textbf{bottom}) against the pre-defined ratio approach (\textbf{top}).}
\label{fig:allocation}

\end{figure}
\textbf{Static-Dynamic Allocation.}
Our adaptive allocation determines the relative static-dynamic composition under a shared global budget rather than enforcing a predefined ratio. As illustrated in Fig. \ref{fig:allocation}, this approach assigns low-motion elements such as walls, stools, and cabinets to the compact static branch while promoting regions with more substantial motion, such as hands and moving dogs, to the dynamic branch. The necessity of this module is quantitatively confirmed in Tab. \ref{t6}, whose removal leads to a \textbf{-0.18 dB} drop in PSNR and a slight model size increase of \textbf{1.7 MB}, underscoring its role in both compactness and fidelity.

\begin{figure}[t]
\centering
\includegraphics[width=\linewidth]{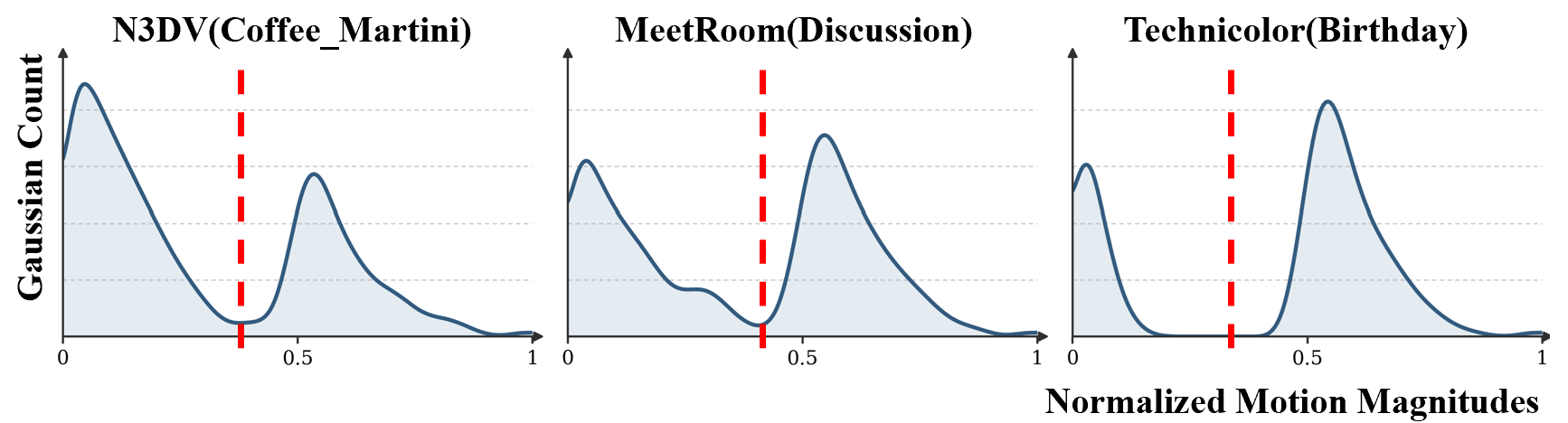}
\caption{Normalized motion distributions and automatically selected static-dynamic thresholds for N3DV (\textit{coffee\_martini}), MeetRoom (\textit{discussion}), and Technicolor (\textit{birthday}), from left to right. The horizontal axis denotes the normalized motion statistic, the vertical axis denotes the Gaussian count in logarithmic scale, and red dashed lines indicate the selected thresholds.}
\label{fig:motion_distribution}
\end{figure}

Fig.~\ref{fig:motion_distribution} further visualizes the normalized motion distributions and automatically selected thresholds on representative sequences from all three datasets. Despite differences in peak ratios, distribution widths, and intermediate-motion populations, the adaptive thresholding criterion produces meaningful thresholds across different motion distributions.

\textbf{Three-phase Training Strategy.} 
We remove the first and third training stages, respectively, to isolate their contributions. As summarized in Tab. \ref{t7}, the initialization stage provides a reliable Gaussian translation prior, which is critical for an accurate static-dynamic allocation. Ablating this stage leads to a sub-optimal allocation that hinders effective training, resulting in a performance drop of \textbf{1.24 dB}. Conversely, the fine-tuning stage further optimizes the Gaussian distributions and attributes after the target count is reached, leading to a notable refinement in reconstruction quality. Its removal causes a drop of \textbf{0.28 dB}, confirming its role in enhancing the final rendering fidelity.

\textbf{Compression Strategy.}
In Tab.~\ref{t7}, \textit{w/o Compression} denotes the fully uncompressed representation, which achieves 32.21 dB at 98.3 MB, compared with 32.14 dB at 31.5 MB after the complete pipeline. Thus, post-training compression reduces storage by approximately $3.12\times$ with a compression-induced quality change of only 0.07 dB. The \textit{w/o Static Compression} and \textit{w/o Dynamic Compression} settings are partial ablations: the other branch remains compressed, so their remaining quality gap relative to the fully uncompressed model is caused by retained lossy compression rather than interaction with training.

\begin{figure}[t]
\centering
\includegraphics[width=\linewidth]{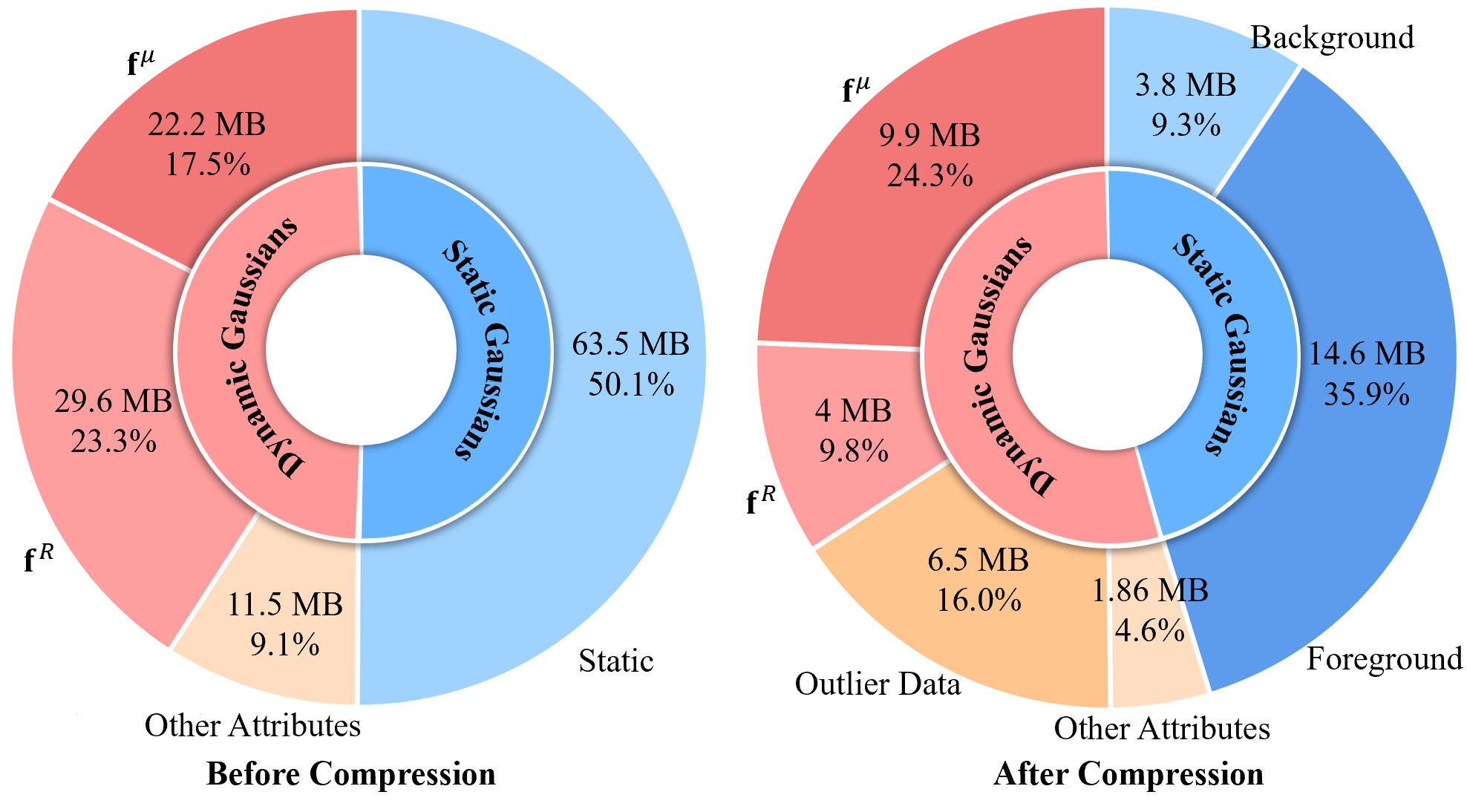}

\caption{Comparison of storage distribution before (\textbf{left}) and after (\textbf{right}) applying our dual-mode hybrid compression on the\textit{ flame\_salmon} sequence. The visualization highlights the significant reduction in model size and the explicit isolation of background and outlier components in the compressed representation.}
\label{fig:compression_pie}
\end{figure}

Fig. \ref{fig:compression_pie} visualizes the storage distribution before and after compression. The regularization loss refines the parameter distribution for subsequent coding, and removing it leads to a \textbf{0.75 dB} drop. Separating outliers from both branches is also important for narrowing the coding range; omitting this step produces a \textbf{2.4 dB} degradation. The small quality change of the complete pipeline mainly originates from attribute quantization and lossy coding, which discard part of the original parameter precision.

\begin{table}[]
\centering
\setlength{\tabcolsep}{8pt} 
\renewcommand{\arraystretch}{1.15}
\caption{Evaluation results of our three-phase training strategy and dual-mode hybrid compression.}
\label{t7}

\scalebox{1}{
\begin{tabular}{c|cc}
\toprule[1pt]
                        & PSNR(dB)$\uparrow$ & Size(MB)$\downarrow$ \\ \hline
w/o Initialization         &  30.90    &   31.8        \\
w/o Fine-tuning  &  31.86    &   31.5        \\
w/o Compression         &  32.21    &   98.3        \\
w/o Static Compression  &  32.19    &   65.5        \\
w/o Dynamic Compression &  32.16    &   64.3        \\
w/o Regularization Loss            &  31.39    &  31.2         \\
w/o Outlier Separation & 29.74     &    26.2       \\
Ours Full            & 32.14     &   31.5        \\ \bottomrule[1pt]
\end{tabular}
}

\end{table}

\textbf{Cross-device Efficiency.}
Using the N3DV checkpoints corresponding to the Ours-l configuration, the complete compressed representation is decoded once in 497 ms on the RTX 3090, 662 ms on the iPad M2, and 836 ms on the iPhone A15. The same rendering resolution is used across all three platforms, and the reported per-frame rendering latencies are averages over the evaluated frames. This startup process is excluded from per-frame rendering latency and amortized over subsequent frames. After decoding, CDGS renders at 5.4, 15, and 32 ms per frame on these platforms, respectively, demonstrating consistent efficiency from a desktop GPU to resource-constrained mobile hardware.

\begin{table}[t]
\centering
\setlength{\tabcolsep}{7pt}
\renewcommand{\arraystretch}{1.15}
\caption{Cross-device decoding and rendering efficiency of CDGS on a desktop GPU and mobile platforms. The compressed representation is decoded once before rendering.}
\label{t_mobile}
\begin{tabular}{c|cc}
\toprule[1pt]
Platform & Decoding (ms) & Rendering (ms) \\ \hline
RTX 3090 & 497 & 5.4 \\
iPad M2 & 662 & 15 \\
iPhone A15 & 836 & 32 \\ \bottomrule[1pt]
\end{tabular}
\end{table}

\section{Conclusion}
In this paper, we presented Constrained Dynamic Gaussian Splatting (CDGS), a framework that redefines dynamic scene reconstruction from an unconstrained fitting task to a budget-constrained optimization problem. Departing from heuristic train-then-prune paradigms, CDGS demonstrates that integrating hardware constraints directly into the training loop is essential for achieving optimal fidelity. Through our differentiable budget controller and adaptive static-dynamic allocation, underpinned by a progressive three-phase training strategy, we successfully resolved the complex spatio-temporal resource competition. Furthermore, coupled with a dual-mode hybrid compression scheme, CDGS ensures that limited capacity is intelligently invested in visually critical motion details while minimizing storage footprint. Extensive experiments validate that CDGS achieves superior rate-distortion performance, offering a robust and practical solution for deploying high-fidelity 4D immersive media on resource-constrained edge devices.

\section*{Acknowledgments}
This work is supported by National Natural Science Foundation of China (62571322, 62431015),STCSM (24ZR1432000, 22DZ2229005), 111 plan (BP0719010), and State Key Laboratory of UHD Video and Audio Production and Presentation.

\bibliographystyle{IEEEtran}
\bibliography{reference}

\begin{IEEEbiography}[
{\includegraphics[width=1in,height=1.25in,clip,keepaspectratio]{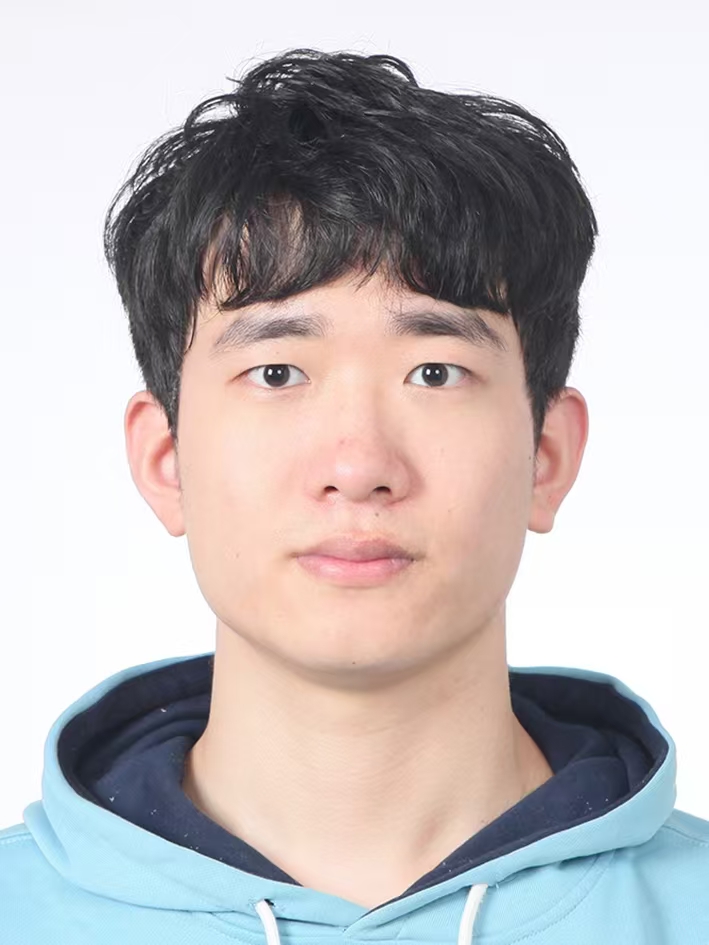}}]{Zihan Zheng} (Student Member, IEEE) received his B.E. degree from Shanghai Jiao Tong University in 2024. He is currently pursuing a Ph.D. in the School of Electronic Information and Electrical Engineering at Shanghai Jiao Tong University. His works have been published in top-tier conferences (e.g.,  CVPR, NeurIPS, ACM-MM and AAAI). His research focuses on volumetric video, 3D reconstruction and 3D generation.
\end{IEEEbiography}

\begin{IEEEbiography}[
{\includegraphics[width=1in,height=1.25in,clip,keepaspectratio]{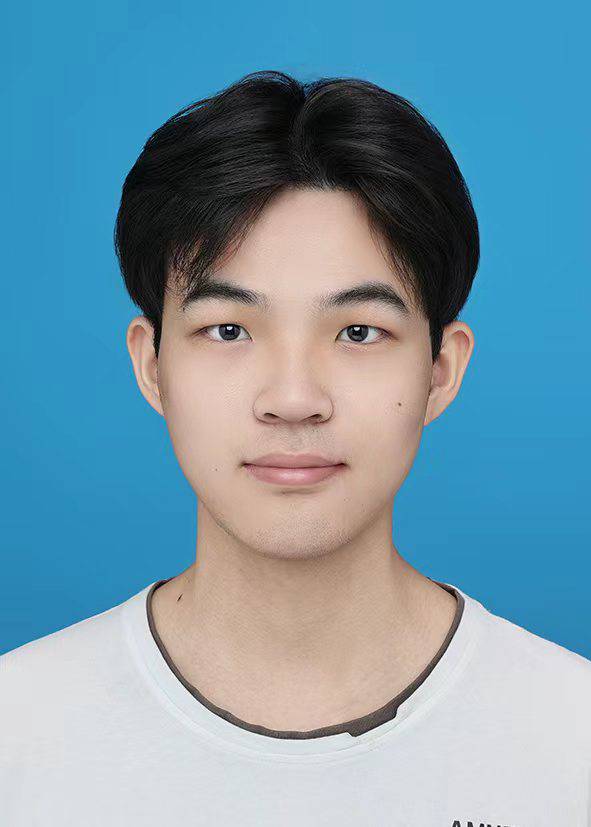}}]{Zhenlong Wu} received his B.E. degree from Shanghai Jiao Tong University in 2025. He is currently pursuing a Ph.D. in the School of Electronic Information and Electrical Engineering at Shanghai Jiao Tong University. His research focuses on diffusion models and 3D reconstruction.
\end{IEEEbiography}

\begin{IEEEbiography}[
{\includegraphics[width=1in,height=1.25in,clip,keepaspectratio]{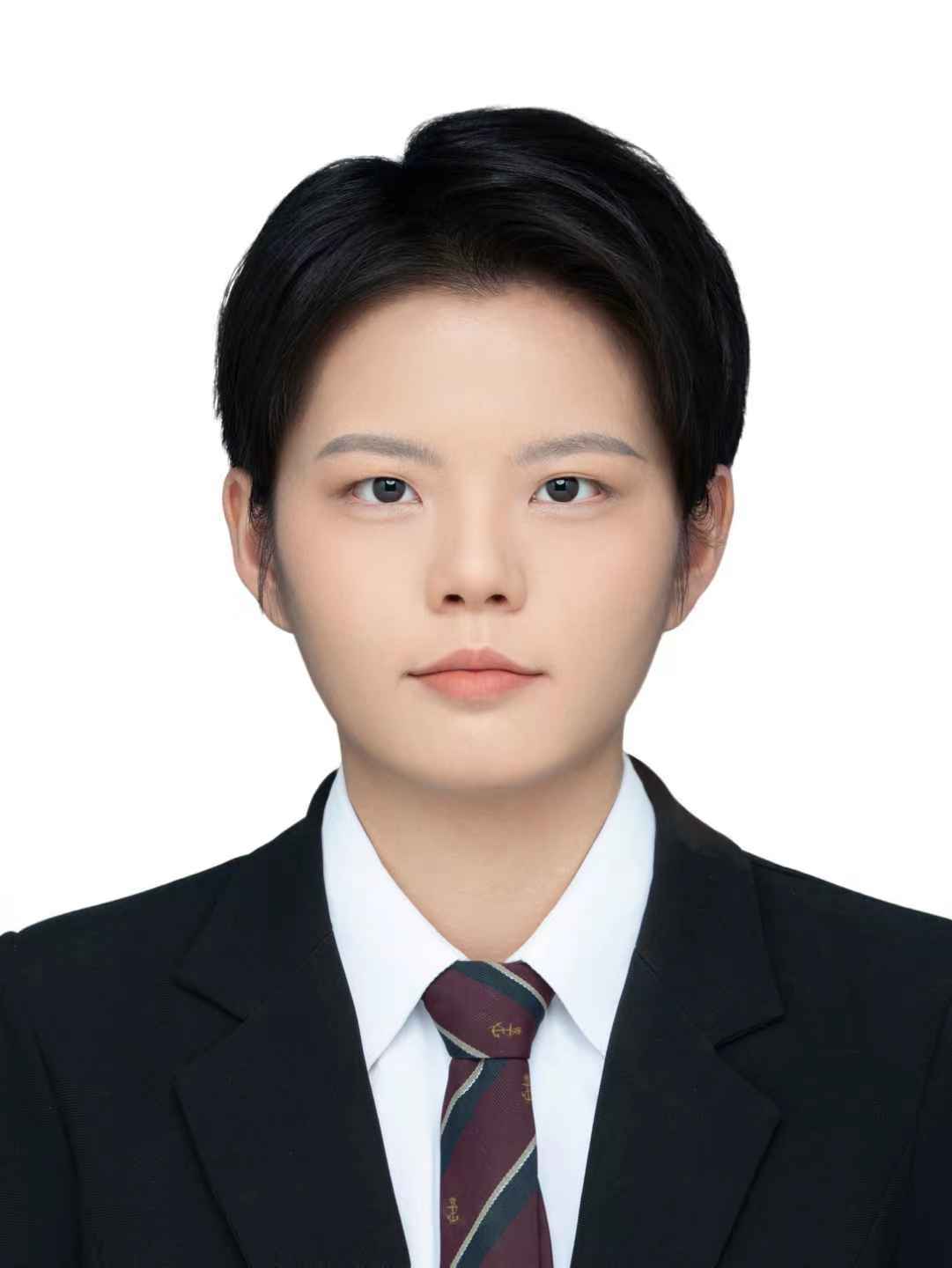}}]{Xuanxuan Wang} received her B.E. degree from Huazhong University of Science and Technology in 2025. She is currently pursuing her Master's degree at the School of Integrated Circuits, Shanghai Jiao Tong University. Her research focuses on volumetric video and 3D reconstruction.
\end{IEEEbiography}

\begin{IEEEbiography}[
{\includegraphics[width=1in,height=1.25in,clip,keepaspectratio]{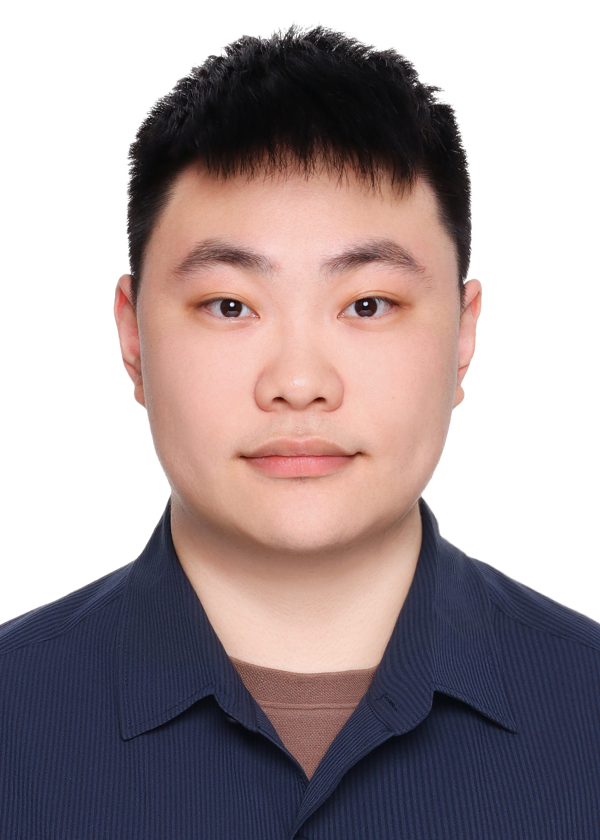}}]{Houqiang Zhong} (Graduate Student Member, IEEE) received his B.E. degree from Shanghai University in 2020 and M.S. degree from ShanghaiTech University in 2023. He is currently pursuing a Ph.D. in the School of Electronic Information and Electrical Engineering at Shanghai Jiao Tong University. His research focuses on volumetric video and 3D reconstruction.
\end{IEEEbiography}

\begin{IEEEbiography}[
{\includegraphics[width=1in,height=1.25in,clip,keepaspectratio]{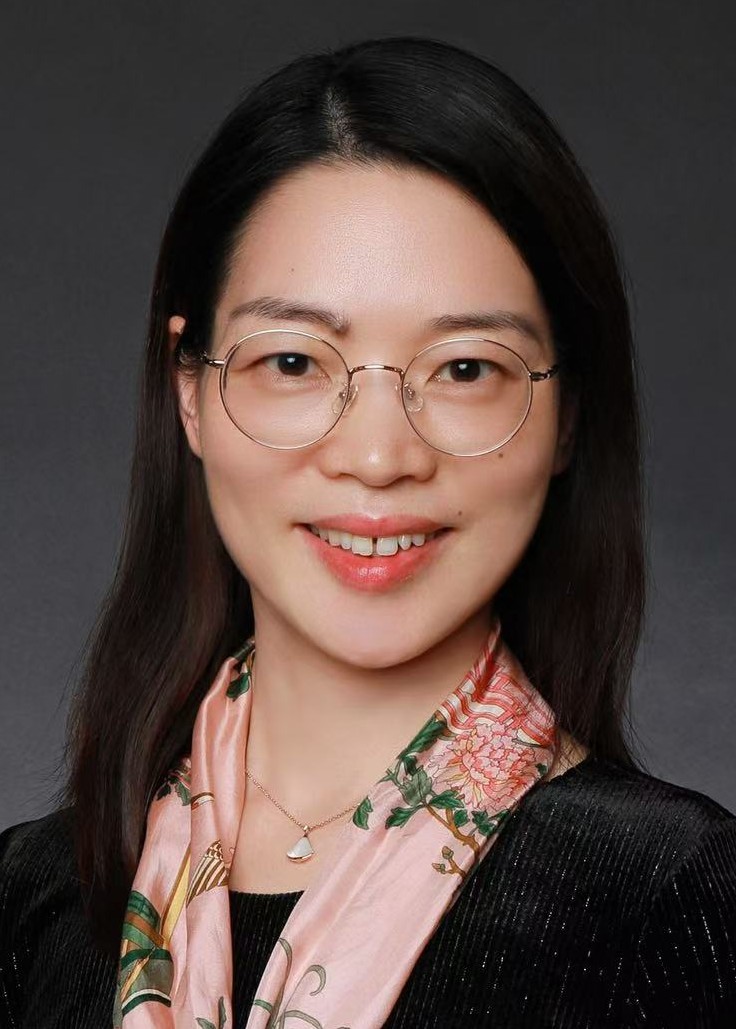}}]{xiaoyun Zhang}(Member, IEEE), Professor of Cooperative Medianet Innovation Center (CMIC), Shanghai Jiao Tong University. She received her Ph.D (supervised by Yuncai Liu) on Pattern Recognition from Shanghai Jiao Tong University and Master (supervised by Zongben Xu) on Applied Mathematics from Xi`an Jiao Tong University respectively. She has co-authorized papers on IEEE TPAMI, TIP, CVPR, ICCV, etc., and her Ph.D. thesis has been nominated as “National 100 Best Ph.D. Theses of China”. She has been a visiting scholar of Harvard University for one year. She is also a member of State Key Laboratory of UHD video and audio production and presentation, and has collaboration with CCTV in video restoration and enhancement.
\end{IEEEbiography}

\begin{IEEEbiography}[
{\includegraphics[width=1in,height=1.25in,clip,keepaspectratio]{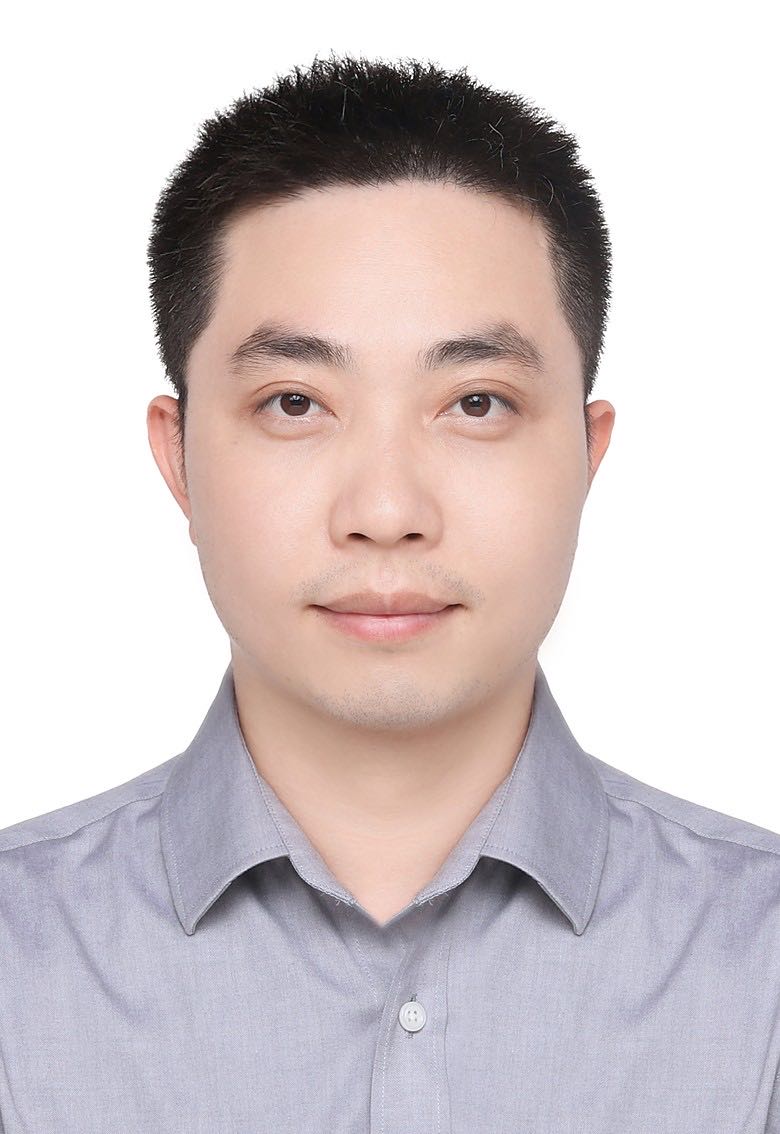}}]{Qiang Hu} (Member, IEEE) received the B.S. degree in electronic engineering from University of Electronic Science and Technology of China in 2013, and the Ph.D. degree in information and communication engineering from Shanghai Jiao Tong University in 2019.  He is currently an Assistant Researcher at Cooperative Medianet Innovation Center, Shanghai Jiao Tong University. Before that, he was an Assistant Researcher at ShanghaiTech University from 2021 to 2023. He was a Postdoc Researcher at ShanghaiTech University from 2019 to 2021. His research interests focus on 2D/3D video compression, 3D reconstruction, and generative intelligence media. His works have been published in top-tier journals and conferences, such as IEEE Transactions on Image Processing (TIP), Conference on Computer Vision and Pattern Recognition (CVPR).
\end{IEEEbiography}

\begin{IEEEbiography}[
{\includegraphics[width=1in,height=1.25in,clip,keepaspectratio]{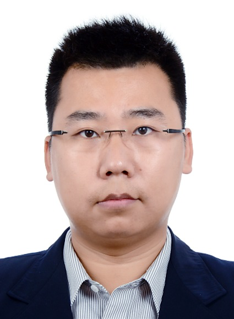}}]{Guangtao Zhai} (Fellow, IEEE) received the B.E. and M.E. degrees from Shandong University, Shandong, China, in 2001 and 2004, respectively, and the Ph.D. degree from Shanghai Jiao Tong University, Shanghai, China, in 2009, where he is currently a Research Professor with the Institute of Image Communication and Information Processing. From 2008 to 2009, he was a Visiting Student with the Department of Electrical and Computer Engineering, McMaster University, Hamilton, ON, Canada, where he was a Post-Doctoral Fellow from 2010 to 2012. From 2012 to 2013, he was a Humboldt Research Fellow with the Institute of Multimedia Communication and Signal Processing, Friedrich Alexander University of Erlangen-Nuremberg, Germany. He received the Award of National Excellent Ph.D. Thesis from the Ministry of Education of China in 2012. His research interests include multimedia signal processing and perceptual signal processing.
\end{IEEEbiography}

\begin{IEEEbiography}[{\includegraphics[width=1in,height=1.25in,clip,keepaspectratio]{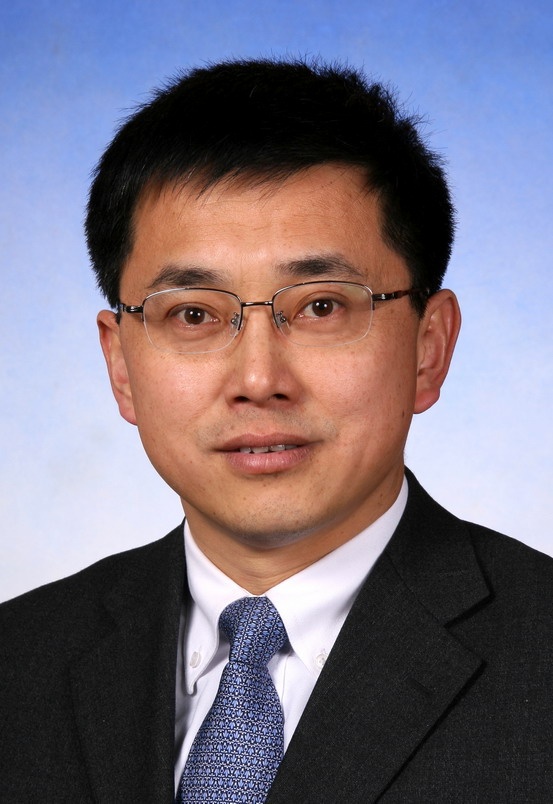}}]{Wenjun Zhang}(Fellow, IEEE) received B.S., M.S., and Ph.D. degrees in electronic engineering from Shanghai Jiao Tong University, Shanghai, China, in 1984, 1987, and 1989, respectively. From 1990 to 1993, he worked as a Postdoctoral Fellow with Philips, Nuremberg, Germany, where he was actively involved in developing the HD-MAC system.He joined the faculty of Shanghai Jiao Tong University in 1993 and became a Full Professor of Electronic Engineering in 1995. He is the Chief Scientist of the Chinese Digital TV Engineering Research Centre, an industry/government consortium in DTV technology research and standardization. His main research interests include digital video coding and transmission, multimedia semantic processing, and intelligent video surveillance.
\end{IEEEbiography}

\end{document}